\documentclass[lettersize,journal]{IEEEtran}
\usepackage{amsmath,amsfonts}
\usepackage{algorithmic}
\usepackage{algorithm}
\usepackage{array}
\pdfoutput=1
\usepackage[caption=false,font=normalsize,labelfont=sf,textfont=sf]{subfig}
\usepackage{textcomp}
\usepackage{stfloats}
\usepackage{url}
\usepackage{verbatim}
\usepackage{graphicx}
\usepackage{cite}
\usepackage{multirow}
\usepackage{booktabs}
\usepackage{multirow}
\usepackage{bm}
\usepackage[table,xcdraw]{xcolor}

\begin{document}

\title{Progressive Inertial Poser: Progressive Real-Time Kinematic Chain Estimation for 3D Full-Body Pose from Three IMU Sensors}

\author{Zunjie Zhu, Yan Zhao, Yihan Hu, Guoxiang Wang, Hai Qiu, Bolun Zheng, Chenggang Yan, Feng Xu
\thanks{This paper was produced by the IEEE Publication Technology Group. They are in Piscataway, NJ.}
\thanks{Manuscript received xx xx, xxxx; revised xx xx, xxxx. (Corresponding author: Bolun Zheng (e-mail: blzheng@hdu.edu.cn).)}

\thanks{Zunjie Zhu and Yihan Hu are with the School of Communication Engineering, Hangzhou Dianzi University, Hangzhou 323000, China. Zunjie Zhu is also with the Key Laboratory of Micro-nano Sensing and IoT of Wenzhou, Wenzhou Institute of Hangzhou Dianzi University, Wenzhou, 325038, China.}
\thanks{Yan Zhao is with the School of Control Science and Engineering, Tiangong University, Tianjin 300387, China.}
\thanks{Guoxiang Wang is with the College of Business, Lishui University, Lishui 310018, China.}
\thanks{Hai Qiu is with Costar Intelligent Optoelectronics Technology Co.,Ltd, China.}
\thanks{Bolun Zheng and Chenggang Yan are with the School of Automation, Hangzhou Dianzi University, Hangzhou 323000, China. Chenggang Yan is also with the Faculty of Applied Sciences, Macao Polytechnic University, Macao 999078, China.}
\thanks{Feng Xu is with the School of software and BNRist, Tsinghua University, Beijing 100084, China.}
}

\markboth{Journal of \LaTeX\ Class Files,~Vol.~14, No.~8, August~2021}%
{Shell \MakeLowercase{\textit{et al.}}: A Sample Article Using IEEEtran.cls for IEEE Journals}

\IEEEpubid{0000--0000/00\$00.00~\copyright~2021 IEEE}

\maketitle

\begin{abstract}
The motion capture system that supports full-body virtual representation is of key significance for virtual reality. Compared to vision-based systems, full-body pose estimation from sparse tracking signals is not limited by environmental conditions or recording range. However, previous works either face the challenge of wearing additional sensors on the pelvis and lower-body or rely on external visual sensors to obtain global positions of key joints. To improve the practicality of the technology for virtual reality applications, we estimate full-body poses using only inertial data obtained from three Inertial Measurement Unit (IMU) sensors worn on the head and wrists, thereby reducing the complexity of the hardware system. In this work, we propose a method called Progressive Inertial Poser (ProgIP) for human pose estimation, which combines neural network estimation with a human dynamics model, considers the hierarchical structure of the kinematic chain, and employs a multi-stage progressive network estimation with increased depth to reconstruct full-body motion in real time. The encoder combines Transformer Encoder and bidirectional LSTM (TE-biLSTM) to flexibly capture the temporal dependencies of the inertial sequence, while the decoder based on multi-layer perceptrons (MLPs) transforms high-dimensional features and accurately projects them onto Skinned Multi-Person Linear (SMPL) model parameters. Quantitative and qualitative experimental results on multiple public datasets show that our method outperforms state-of-the-art methods with the same inputs, and is comparable to recent works using six IMU sensors.
\end{abstract}


\section{Introduction}
\IEEEPARstart{V}{irtual} reality technology offers users an immersive experience through computer-generated environments, with precise full-body motion tracking playing a crucial role in enhancing this experience. The innovative integration of virtual reality and motion capture ensures a seamless alignment between real-world motions and virtual scenes, and opens up new interactive possibilities for various fields such as motion analysis \cite{ref1} and healthcare applications \cite{ref2}.
\IEEEpubidadjcol

In current virtual reality applications, one of the mature high-precision motion capture solutions is the vision-based method \cite{ref3,ref4}. This method estimates human pose using multiple RGB cameras with or without markers \cite{ref5}, but it is prone to being affected by external environments and application scenarios. Wearable inertial sensors also provide a satisfactory solution for motion capture, overcoming the inherent issues of occlusions and limited monitoring areas in vision \cite{ref6, ref35}. For example, the commercial inertial motion capture system Xsens \cite{ref7} obtains motion information about human joints from 17 or more inertial sensors. In recent years, research has further reduced the required sensor data to six, which are sparsely worn on the head, pelvis, wrists, and ankles, and uses sparse inertial sensor data to estimate 3D human pose in real time \cite{ref8,ref9,ref10}. However, additional devices worn on the lower-body limit motion diversity and personal comfort. Therefore, a head-mounted display (HMD) and two handheld controllers are usually used for interactions in typical virtual reality settings \cite{ref11,ref12}.

To reduce the number of devices and improve portability in applications such as virtual reality, we aim to improve the applicability and efficiency of full-body pose estimation using only the acceleration and rotation provided by three pure inertial sensors worn on the head and wrists. It is a challenging inverse kinematics (IK) problem to directly estimate full-body joint poses based on known inertial constraints without position knowledge of sparse upper-body joints. However, traditional IK methods neglect the human dynamics constraints, causing joint rotation errors to accumulate along the kinematic chain and result in unnatural deformation of the end-effector \cite{ref36}. We observe significant motion correlation between adjacent joints and introduce a local region modeling strategy, which progressively estimates joint poses with the same or similar depth in the corresponding region according to the order of the kinematic chain depth increase in multiple stages. The rotation of ancestor joints should be estimated earlier than the rotation of descendant joints, because joints with smaller depths are closer to the center of the body and influence all joints at their subsequent depths, thus determining the posture of the entire skeleton \cite{ref14}. This estimation strategy effectively reduces error accumulation and improves the accuracy and naturalness of virtual full-body character reconstruction.

\begin{figure*}[!t]
	\centering
	\includegraphics[width=7.1in]{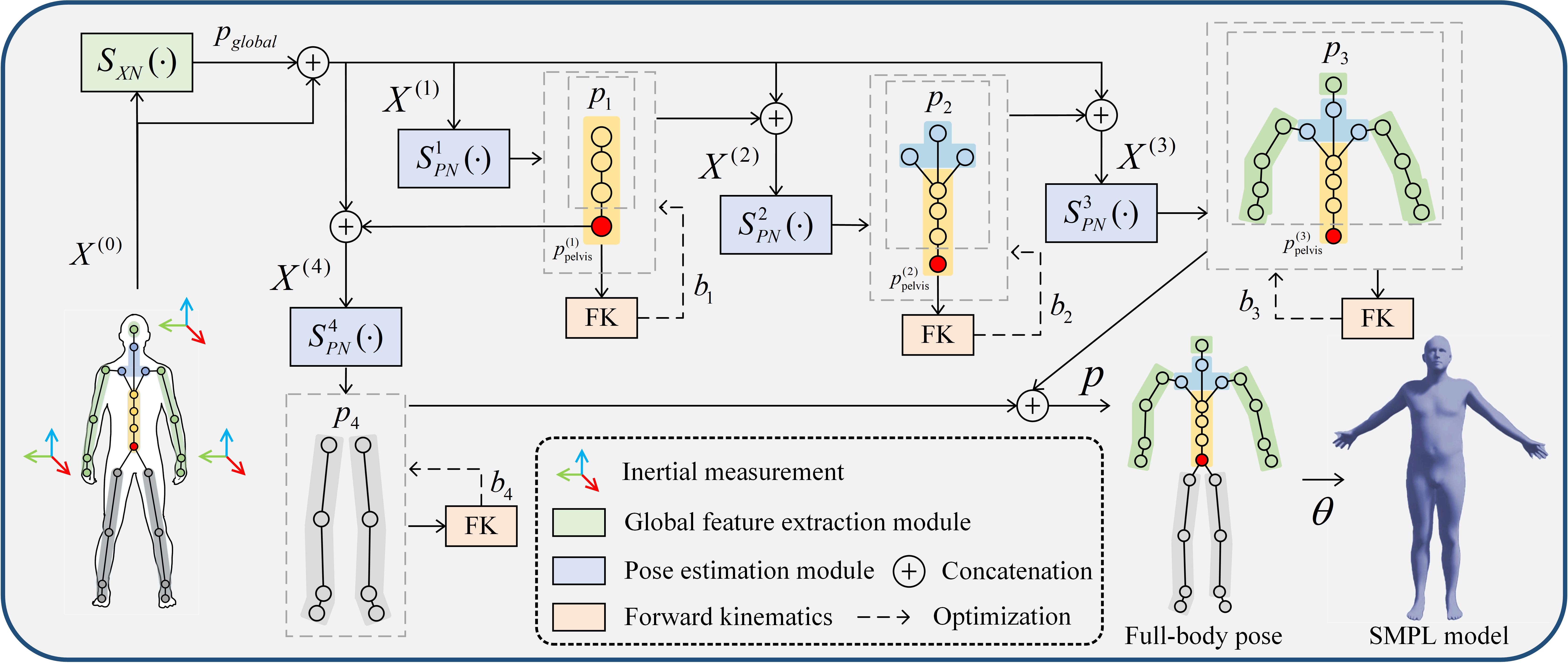}
	\caption{The pipeline of our method. We divide the human body into four regions based on the hierarchical structure of the kinematic chain and use multi-stage progressive pose estimation to achieve real-time full-body motion synthesis. First, the full-body motion information ${\bm{p}_{global}}$ is roughly estimated from the IMU measurements, and its output is combined with the IMU measurements $\bm{X}^{(0)}$ as $\bm{X}^{(1)}$. The progressive estimation process is divided into four stages:
		(1) The first stage estimates joint poses in the first region from input $\bm{X}^{(1)}$, and the output $\left[ {\bm{p}_{1}},\bm{p}_{{\text{pelvis}}}^{(1)} \right]$ is concatenated with $\bm{X}^{(1)}$ to form $\bm{X}^{(2)}$;
		(2) The second stage estimates joint poses in the second region from input $\bm{X}^{(2)}$, and output $\left[ {\bm{p}_{2}},\bm{p}_{{\text{pelvis}}}^{(2)} \right]$ is concatenated with $\bm{X}^{(1)}$ to form $\bm{X}^{(3)}$;
		(3) The third stage estimates joint poses in the third region from input $\bm{X}^{(3)}$, and outputs $\left[ {\bm{p}_{3}},\bm{p}_{{\text{pelvis}}}^{(3)} \right]$ represents the pose of the upper-body joints including the pelvis;
		(4) The fourth stage estimates joint poses in the fourth region from concatenated input of $\bm{X}^{(1)}$ and $\bm{p}_{{\text{pelvis}}}^{(1)}$, and outputs $\bm{p}_{4}$ represents lower-body poses.
		Finally, we combine $\left[ {\bm{p}_{\text{3}}},\bm{p}_{{\text{pelvis}}}^{(3)} \right]$ and $\bm{p}_{\text{4}}$ to obtain full-body poses and project them onto the SMPL model.}
	\label{fig1}
\end{figure*}

Consequently, to achieve realistic real-time full-body motion synthesis, we propose a kinematic chain estimation method called Progressive Inertial Poser (ProgIP), which progressively estimates joint poses along the depth of the kinematic chain using only the acceleration and rotation measurements provided by three IMU sensors worn on the head and wrists, as shown in Fig. \ref{fig1}. The well-designed TE-biLSTM encoder provides both global and local understanding of the inertial signals, enhancing the quality of motion reconstruction in online mode. The MLP-based decoder shares high-dimensional complex features from the encoder to project and transform the pose features into the SMPL model parameters. To the best of our knowledge, there is currently no task specifically designed to estimate full-body poses using only three pure IMU sensors from the head and wrists. We demonstrate the effectiveness of ProgIP on challenging public datasets (including AMASS, DIP-IMU, and TotalCapture), achieving state-of-the-art performance for full-body pose estimation with three sets of inertial inputs, and generating realistic real-time animated demonstrations within an acceptable delay.

The contributions are summarized as follows:
\begin{itemize}
	\item{}We propose ProgIP, which uses only three IMU sensors from the head and wrists to guide the regression of full-body joint rotation. ProgIP progressively estimates joint poses in four regions according to the depth of the kinematic chain, where the TE-biLSTM encoder and the MLP-based decoder focus on the dependencies between adjacent joints by leveraging both local and global information from the inertial data. Additionally, we incorporate joint position consistency loss calculated by forward kinematics into the iterative optimization, which effectively reduces the accumulation of rotation errors in the kinematic chain. ProgIP offers a reference scheme for full-body motion capture with only available inertial tracking inputs from the head and wrists.
	
	\item{}We present live demonstrations that capture a variety of challenging motions while allowing performers to move freely. ProgIP generates realistic, smooth motion and achieves real-time inference speeds, making it suitable for online applications in virtual reality environments.
\end{itemize}

\section{Related works}
Motion capture focusing on full-body digitization has been extensively studied in academia. Existing vision- and marker-based works have achieved numerous remarkable results. For example, commercial motion capture systems such as Vicon \cite{ref15} and OptiTrack \cite{ref16} provide high-quality solutions for the gaming and film industries. Our method requires only three sparse IMU sensors worn on the head and wrists as input sources, so in this section, we mainly review closely related solutions for estimating full-body poses using wearable sensors, including 6 DOF (rotation and translation) inputs from the head and wrists, and pure inertial (acceleration and rotation) inputs from sparse joints.
\subsection{6 DOF Inputs From The Head And Wrists}
Some recent research aims to address the challenge of generating full-body poses from input in realistic virtual reality settings, relying on 6 DOF information accessible from an HMD and two handheld controllers to track human motion in real-time and generate realistic motion while observing specific body parts.
Jiang et al. \cite{ref12} were the first to propose a learning-based method to estimate full-body poses using only the rotation and translation inputs from the user's head and hands, called AvatarPoser. This method uses holistic avatar representation to overcome the limitations of floating avatars in virtual reality interactions.
On this basis, Du et al. \cite{ref17} proposed AGRoL to address the challenging lower-body motion and generating smoothness, and the designed diffusion model with sparse tracking conditions reconstructed the full-body motion from 6 DOF tracking information for the first time.
Aliakbarian et al. \cite{ref18} also used a generative model to learn the conditional distribution of full-body poses based on the knowledge from the head and hands, named FLAG, and proposed an optimized pose prior and a new approach based on conditional normalized flow to generate high-quality poses.
In order to reduce the influence of the visual range of the HMD on hand interaction, Streli et al. \cite{ref19} proposed HOOV, which supplements the headset information with continuous signals from a wristband to estimate the current hand position outside the visible field of view.
Some other studies suggest adding additional signal sources to track pelvic motion. Yang et al. \cite{ref20} estimated lower-body poses based on tracking signals from the head, hands, and pelvis, known as LoBSTr, which employs velocity to represent the correlation between the upper-body signal and the lower-body motion and finally obtained the full-body virtual animation through the IK solver.

\subsection{Pure Inertial Inputs From Sparse Joints}
To overcome the limitations of system cost and dense placement, previous works have worn sparse pure IMU sensors on different body parts for motion tracking to accurately estimate full-body poses. Early work \cite{ref21} attempted to use only five sparse accelerometers to continuously match the collected data with the closest data in the existing database for motion capture.
Recently, the groundbreaking work for full-body pose estimation using inertial sensors that measure acceleration and rotation simultaneously is SIP, proposed by Marcard et al. \cite{ref8}. SIP can optimize all poses in the sequence at once, but it does not meet real-time requirements.
Therefore, Huang et al. \cite{ref9} proposed to use deep learning to learn temporal pose prior, called DIP, which is the first deep learning method based on a bidirectional RNN to estimate human pose and deploy sliding window architecture to maintain real-time capabilities.
Yi et al. believed that directly regressing rotation from sparse IMU sensors is extremely challenging, so they proposed TransPose \cite{ref10} to estimate joint positions as the intermediate representation of estimated joint relative rotations, and suggested developing the pose estimation task in a multi-stage manner.
On this basis, PIP \cite{ref22} proposed a physics-aware motion optimizer to refine motion to satisfy physical constraints, which is a significant improvement over previous technologies.
Aiming to address the challenges of inconsistent prediction time and joint motion drift, Jiang et al. \cite{ref23} proposed TIP, which uses the Transformer to improve the reasoning ability by explicitly taking its past predictions as input and achieves real-time enhanced reconstruction of full-body motion using only six IMU sensors.
Zhang et al. \cite{ref24} proposed a part-based human pose estimation model focusing on the spatial relationship between human body parts and IMU sensors. Unlike previous work that used only temporal information to reconstruct complex motions, the proposed model focuses on the exclusive features of corresponding body regions to improve the estimation accuracy.
Mollyn et al. \cite{ref25} explored using built-in IMU sensors from low-cost consumer products to guess the optimal joint poses, called IMUPoser, which builds an intriguing real-time ecosystem to automatically track available equipment without additional external facilities, so that it is particularly suitable for applications in the healthcare market.

In summary, all the methods reviewed in this section either require additional joint position information, additional tracking inputs from more than three joints, or face difficulties in predicting accurate full-body poses in real time from sparse inputs. Our proposed ProgIP can effectively estimate full-body poses using only pure inertial inputs from three IMU sensors worn on the head and wrists. It performs progressive pose estimation along the depth of the kinematic chain and employs a straightforward network structure based on Transformers and RNNs. Through the analysis and comparison of these existing methods, we aim to develop a simple, practical, and cost-effective full-body pose estimation technique to advance the development and application of motion capture solutions.

\section{Method}
\subsection{Problem Formulation}
\begin{figure}[!b]
	\centering
	\includegraphics[width=3.1in]{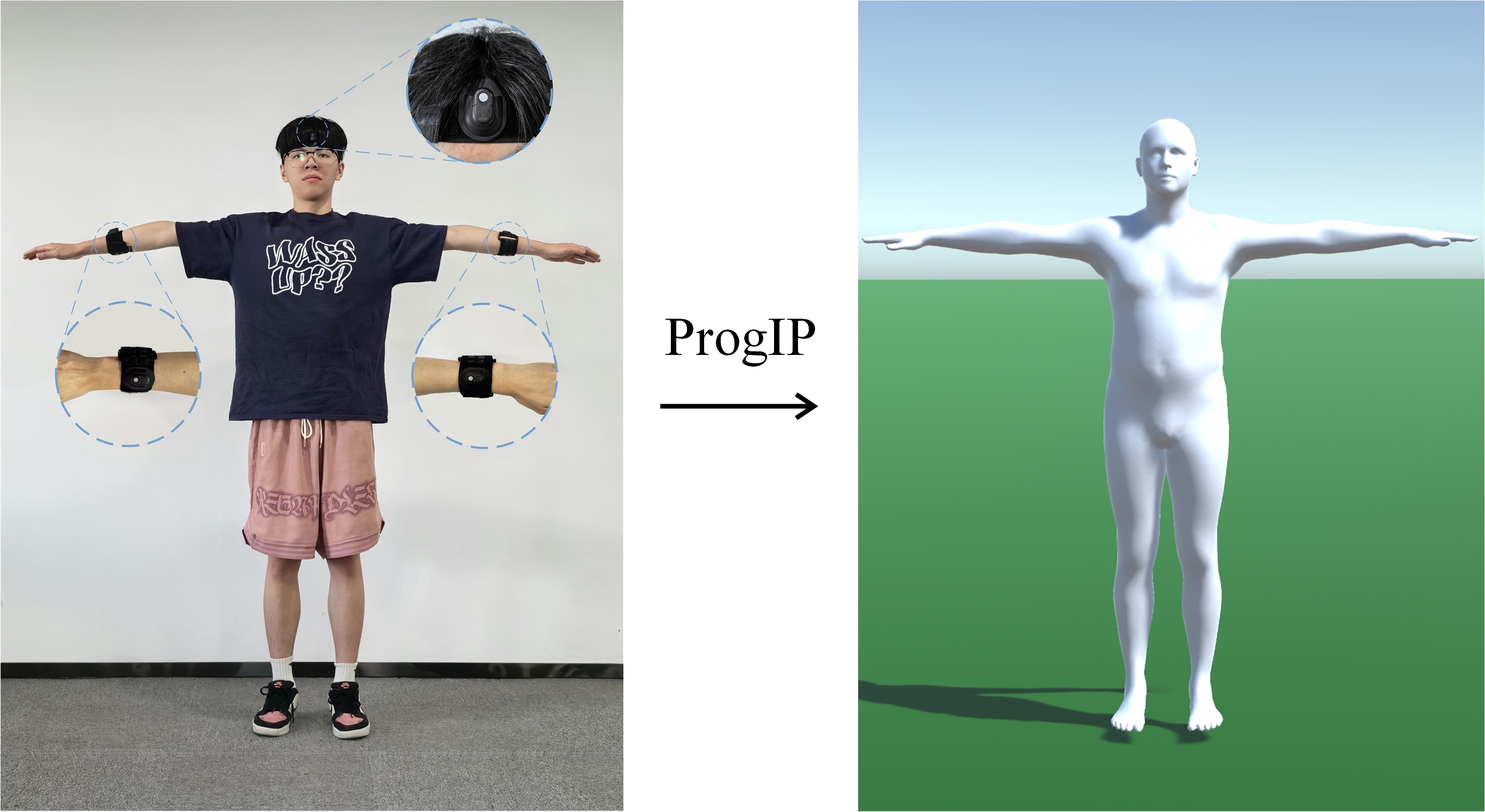}
	\caption{The proposed ProgIP generates full-body poses by using only acceleration and rotation data from the head and wrists. The left image illustrates the IMU placement, where the sensors are tightly bound with arbitrary orientations.}
	\label{fig2}
\end{figure}
We introduce a full-body pose estimation method that aims to reconstruct human motion in real time from continuous inertial measurements collected by a set of sparse IMU sensors worn on the head and wrists, as shown in Fig. \ref{fig2}. 
This severely under-constrained problem is challenging, for which
we aim to estimate full-body poses ${{P^{1:J}}}=\left\{ {\bm{M}_{j,t}}(\theta ) \right\}\in {{\mathbb{R}}^{J\times N}}$ through a learning-based approach using the observed sparse joint feature sequences ${F^{1:S}} = \left\{ {\left( {{\bm{a}_{s,t}},{\bm{R}_{s,t}}} \right)} \right\}_{s = 1}^S \in {\mathbb{R}^{S \times \left( {A + C} \right)}}$ and the state-of-the-art human parameterization model SMPL \cite{ref26}, where $J$ is the number of joints in the full-body skeleton, $N$ is the dimension of the output joint poses, $S$ is  the number of joints tracked by IMU sensors, $A$ and $C$ represent the dimensions of acceleration and direction, respectively. $\left( {{\bm{a}_{s,t}},{\bm{R}_{s,t}}} \right) \in {\mathbb{R}^{A + C}}$ represents a set of acceleration and rotation measurements from the $s$-th IMU sensor at the $t$-th frame. The SMPL model is denoted by ${\bm{M}_{j,t}}\left( \theta  \right)$, where $\theta$ is the pose parameter. We omit the shape parameter.

\subsection{Input and Output Representation}
We use the rotation and acceleration measurements of each IMU as the raw inputs for the system. We aligned these measurements into the same reference frame and scaled the acceleration 30 times to be suitable for neural network. Referring to \cite{ref12,ref17}, we use the rotation ${\bm{R}_{t}}$ to calculate the angular velocity ${{\bm{W}}_{t}}=\bm{R}_{t-1}^{-1}{\bm{R}_{t}}$, which provides additional dynamic information. Since the continuous 6D rotation representation is suitable for neural network training, we discard the last column of the rotation matrix to obtain the 6D rotation representation \cite{ref27}. Therefore, the final input representation $\bm{X} = \left[ {\left\{ {{\bm{a}_1},{\bm{R}_1},{\bm{W} _1}} \right\}, \cdots ,\left\{ {{\bm{a}_S},{\bm{R}_S},{\bm{W} _S}} \right\}} \right] \in {\mathbb{R}^{S \times 15}}$ is a concatenated vector of acceleration, rotation, and angular velocity from all given sparse IMU sensors.
We set the number of worn IMU sensors to $S=3$, and the input feature dimension at each time step is 45. The output is the global rotation of the pelvis and the local rotation of other joints relative to the parent joints, represented by 6D rotation. According to relevant studies \cite{ref10}, we do not assign rotational degrees of freedom to wrists, hands, ankles and feet because there are no observations to resolve these, so the output feature dimension of each time step is 96.

\subsection{Backbone Network}
\begin{figure}[!tb]
	\centering
	\includegraphics[width=3.4in]{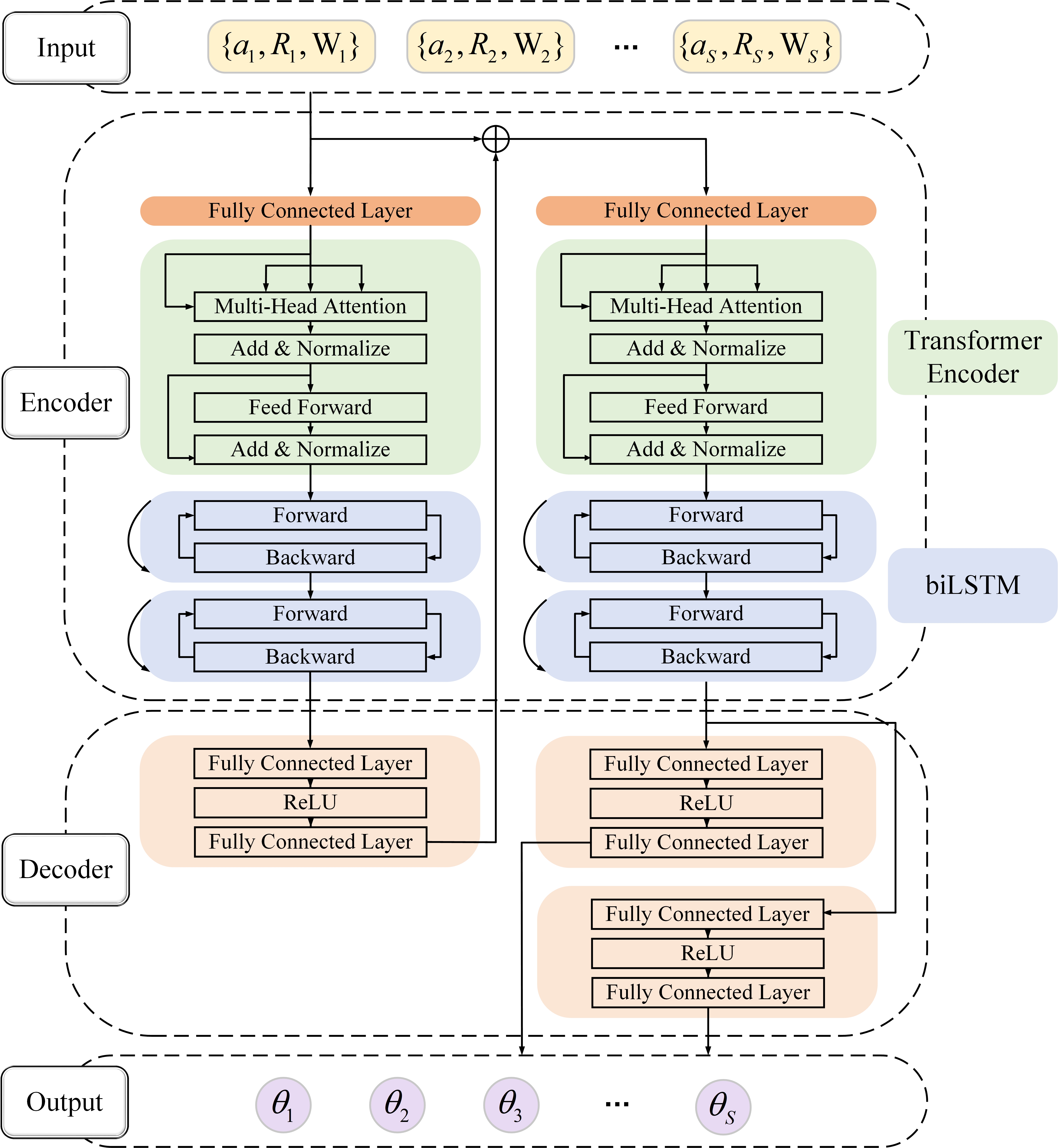}
	\caption{The detailed structure of the backbone network in the pipeline. It mainly includes the TE-biLSTM encoder and the MLP-based decoder, and the final two decoders in the network output the pose of the pelvic joint and the poses of the other joints, respectively.}
	\label{fig3}
\end{figure}

We hereby introduce the detailed structure of the backbone network in the proposed ProgIP as shown in Fig. \ref{fig3}. It primarily consists of two parts: the encoder and the decoder.

The encoder is composed of three main components: a single-layer fully connected (FC) layer, a Transformer Encoder, and a two-layer bidirectional long short-term memory (biLSTM) network. The main purpose of the FC layer is to process and transform the input information and project the input data into a high-dimensional space. The biLSTM layer is added to the Transformer Encoder layer to jointly extract the temporal features of inertial sequences, which leverages the parallelism of the self-attention mechanism in the Transformer Encoder and the memory of the gating mechanism in the biLSTM to provide an understanding of global and local information \cite{ref28}. The well-designed encoder enhances the performance in maintaining the temporal continuity of human motion and real-time application tasks, and can better adapt to the inertial input in practical scenarios.

The decoder is a two-layer MLP structure that shares high-dimensional pose features from the encoder, and the ReLU activation function is applied to enhance the nonlinear capability of the network. We set the global rotation decoder and the pose feature decoder according to the different functions performed, in which the global rotation decoder generates the global direction represented by the pelvic rotation to guide the navigation of the character.

For a given input signal, we apply a linear projection in the FC layer to expand the features to 256 dimensions, and then feed the feature output generated by the Transformer Encoder into two biLSTM networks with a width of 256 to process the data. We set the number of heads to 8 and the number of self-attention layers to 3. The linear operation of the decoder maps the encoder output to a specific dimension, and applies a ReLU activation function to convert the input vector into an embedding of 256 dimensions. The SMPL pose parameters are finally output represented by 6D rotation. Table \ref{tab1} shows the details of our network architecture.
\begin{table}[!tb]
	\caption{Specific parameters of the backbone network. RNN-Layer denotes the number of layers of biLSTM, TF-Layer represents the number of layers of Transformer Encoder, and Head indicates the number heads in the multi-head attention mechanism.}
	\label{tab1}
	\setlength{\tabcolsep}{1mm}{
		\begin{tabular}{@{}ccccccc@{}}
			\toprule
			\multicolumn{2}{c|}{Structure} & ${S_{XN}}\left(  \cdot  \right)$ & $S_{PN}^1\left(  \cdot  \right)$ & $S_{PN}^2\left(  \cdot  \right)$ & $S_{PN}^3\left(  \cdot  \right)$ & $S_{PN}^4\left(  \cdot  \right)$ \\ \midrule
			\multicolumn{1}{c|}{\multirow{5}{*}{Encoder}} & \multicolumn{1}{c|}{Input}     & 45  & 141 & 165 & 183 & 147 \\
			\multicolumn{1}{c|}{}          & \multicolumn{1}{c|}{FC layer}    & 256 & 256 & 256 & 256 & 256 \\
			\multicolumn{1}{c|}{}                         & \multicolumn{1}{c|}{RNN-layer} & 2   & 2   & 2   & 2   & 2   \\
			\multicolumn{1}{c|}{}                         & \multicolumn{1}{c|}{TF-layer}  & 3   & 3   & 3   & 3   & 3   \\
			\multicolumn{1}{c|}{}                         & \multicolumn{1}{c|}{Head}      & 8   & 8   & 8   & 8   & 8   \\ 
			\midrule
			\multicolumn{1}{c|}{\multirow{2}{*}{Recoder}} & \multicolumn{1}{c|}{Hidden}    & 256 & 256 & 256 & 256 & 256 \\
			\multicolumn{1}{c|}{}                         & \multicolumn{1}{c|}{Output}    & 96  & 24  & 42  & 72  & 24  \\ 
			\bottomrule
	\end{tabular}}
\end{table}

\subsection{Multi-stage Progressive Kinematic Chain Estimation}
Inspired by observations from \cite{ref24,ref29,ref30}, we propose incorporating four regions into the multi-stage estimation task, with each stage featuring a backbone network structure to estimate joint poses within the corresponding region. The effectiveness primarily relies on the following two key observations:
(1) The region-based structure can enhance the interdependence between adjacent joints and mitigate the negative effects of weakly correlated joints, thereby enabling effective learning of the unique joint features in local body regions;
(2) The range of estimated joint poses is progressively expanded with increasing depth, preventing the focus from being limited to local regions and ensuring consideration of the overall structure.
 
\begin{figure}[!b]
	\centering
	\subfloat[]{\includegraphics[width=1.9in]{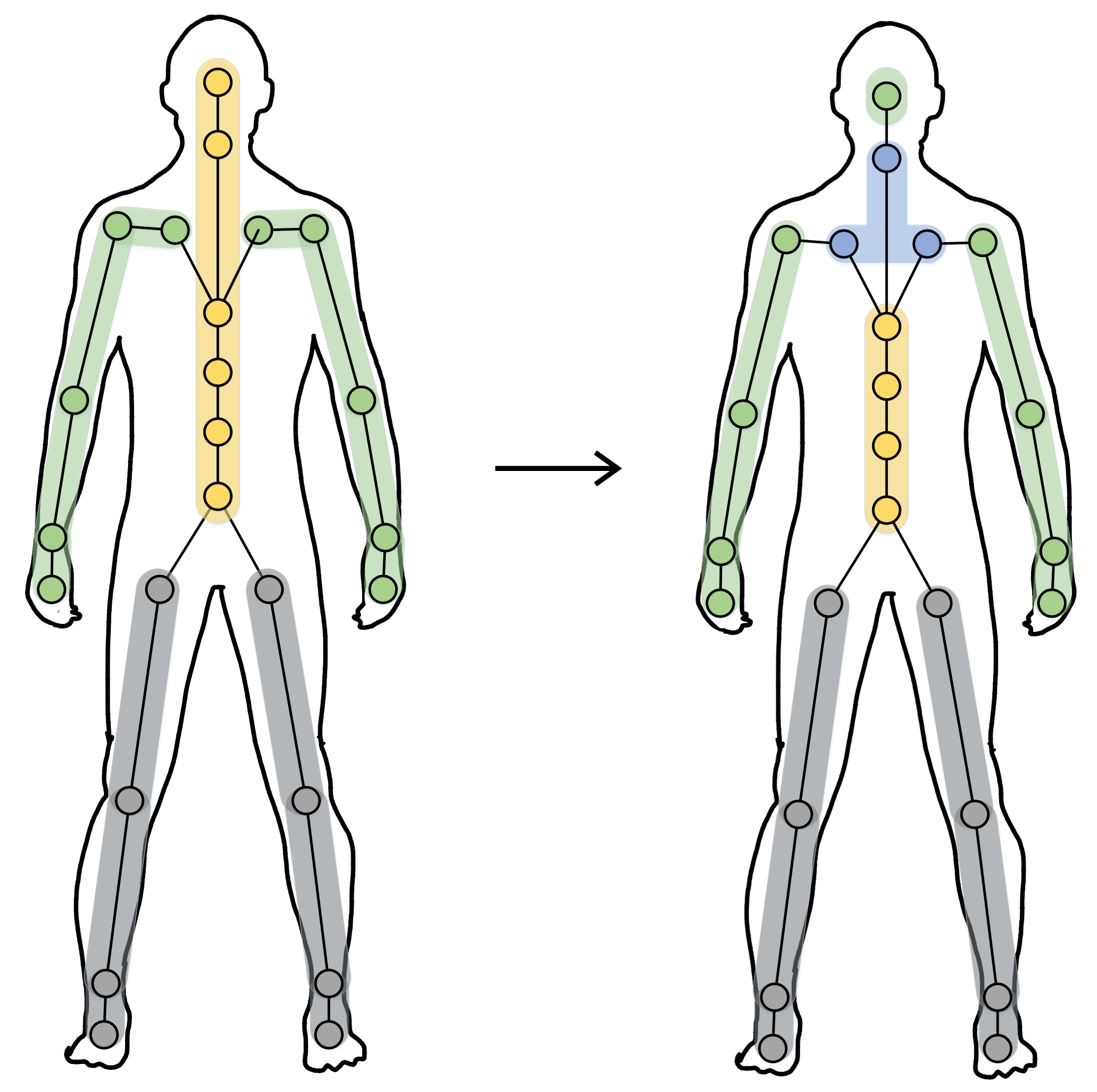}%
		\label{fig4-1}}
	\hfil
	\subfloat[]{\includegraphics[width=1.4in]{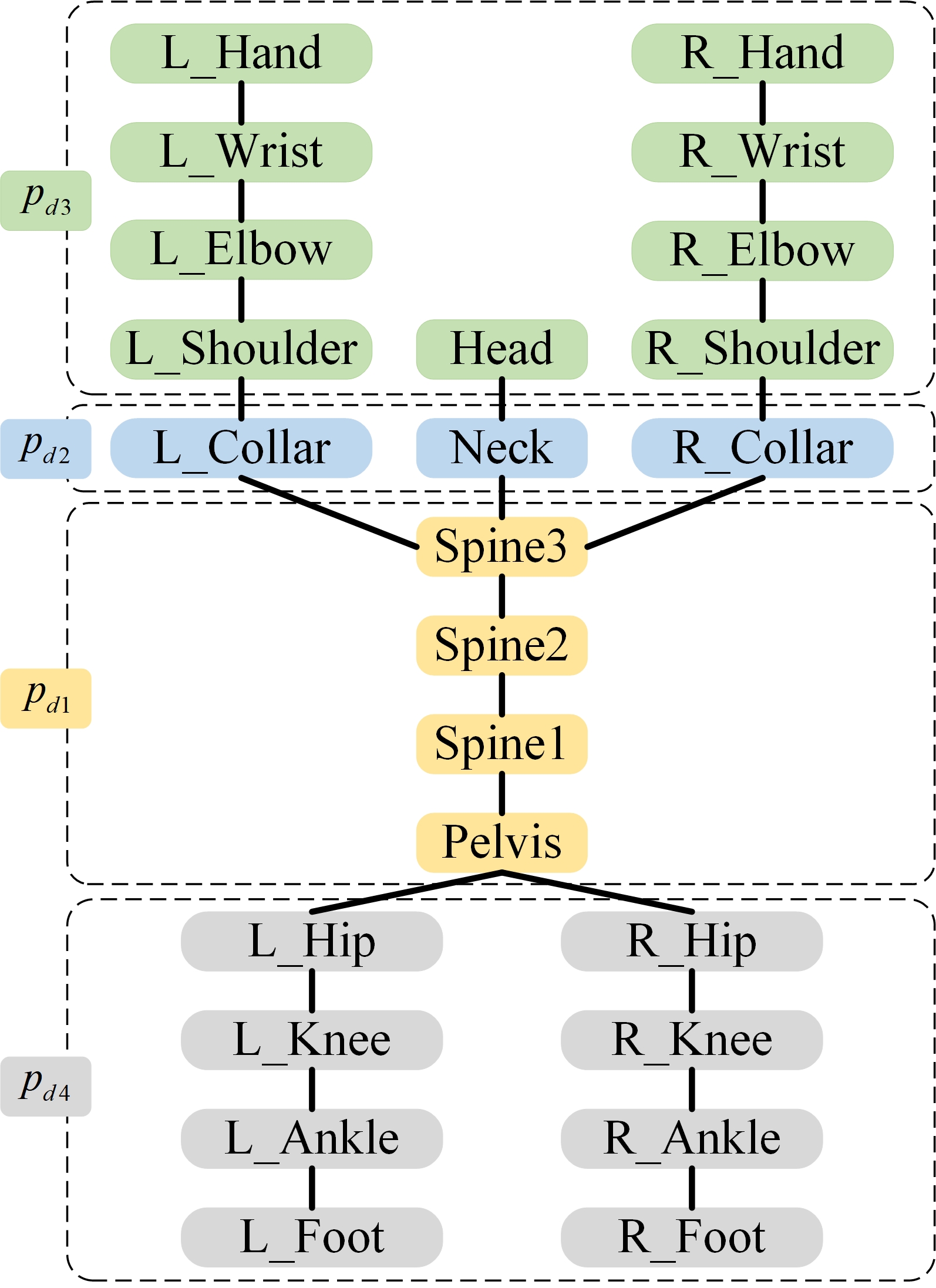}%
		\label{fig4-2}}
	\caption{Human body region division and hierarchical structure in the dynamic model. (a) We modify the human body region division from previous work by dividing the human body into four regions, with specific attention given to the neck, left collar, and right collar as a transition region. (b) We present the joint poses of the four divided body regions in detail in order of kinematic chain depth, gradually increasing from the pelvis to the upper and lower body.}
	\label{fig4}
\end{figure}

Here, we appropriately modify the region division in \cite{ref24} as shown in Fig. \ref{fig4-1}, considering that the neck joint and the collar joints are the transition links between the trunk and the upper-body. Therefore, we pay special attention to these three joints and divide the joints of the full-body into four regions, and the joint poses of each region can be represented as $\bm{p}=\left[ {{\bm{p}_{d1}},{\bm{p}_{d2}},{\bm{p}_{d3}},{\bm{p}_{d4}}} \right]$, where $\bm{p}_{d1} = $ $[$ $ {\bm{p}_{{\text{Pelvis}}},\bm{p}_{{\text{Spine1}}},\bm{p}_{{\text{Spine2}}},\bm{p}_{{\text{Spine3}}}} $ $]$, $\bm{p}_{d2}$ $ = $ $[$ ${{\bm{p}_{{\text{Neck}}}}, {\bm{p}_{{\text{R\_Collar}}}},{\bm{p}_{{\text{L}}\_{\text{Collar}}}}}$ $]$, ${\bm{p}_{d3}} = [ {{\bm{p}_{{\text{Head}}}},{\bm{p}_{{\text{R\_}}{\text{Shoulder}}}},{\bm{p}_{{\text{L\_Shoulder}}}},{\bm{p}_{{\text{R\_Elbow}}}},{\bm{p}_{{\text{L\_Elbow}}}}}]$, ${\bm{p}_{d4}} = $ $ [{{\bm{p}_{{\text{R\_Hip}}}},{\bm{p}_{{\text{L\_Hip}}}},{\bm{p}_{{\text{R\_Knee}}}},{\bm{p}_{{\text{L\_Knee}}}}}]$. The joint poses in each specific body region divided according to the depth order of the kinematic chain are shown in Fig. \ref{fig4-2}.

We design a global feature extraction module ${S_{XN}\left(  \cdot  \right)}$ to roughly estimate full-body poses, which further enhances the overall consistency of global information.
The input of ${S_{XN}\left(  \cdot  \right)}$ consists of inertial measurements from the head and wrists $\bm{X}$, and the output includes the global rotation of the pelvis and the local rotations of other joints relative to the parent joints, expressed as:
\begin{equation}
	{\bm{p}_{global}} = {S_{XN}\left( {\bm{X}} \right)}
\end{equation}

Subsequently, we concatenate the global information ${\bm{p}_{global}}$ output by $S_{XN}\left(  \cdot  \right)$ with the inertial measurements $\bm{X}$ to form the combined input $\left[ {\bm{X},{\bm{p}_{global}}} \right]$. In the subsequent progressive pose estimation task, the inputs $\bm{X^{\left( i \right)}}$ for the four stages are expressed as $\bm{X^{\left( 1 \right)}} = \left[ {\bm{X},{\bm{p}_{global}}} \right]$, $\bm{X^{\left( 2 \right)}} = \left[ {\bm{X^{\left( 1 \right)}},{\bm{p}_{d1}}} \right]$, $\bm{X^{\left( 3 \right)}} = \left[ {\bm{X^{\left( 2 \right)}},{\bm{p}_{d2}}} \right]$, and $\bm{X^{\left( 4 \right)}} = \left[ {\bm{X^{\left( 1 \right)}},\bm{p}_{{\text{pelvis}}}^{\left( 1 \right)}} \right]$.

We use the TE-biLSTM encoder and the MLP-based decoder to extract depth features that combine past and future time information from the input to calculate joint poses. The output of the network for the first three stages includes the global rotation of the pelvis ${\bm{p}_{{\text{pelvis}}}}$ and the local rotation of other joints ${\bm{p}_i}$, and the mapping relationship is expressed as: 
\begin{equation}
	\left[ {\bm{p}_{{\text{pelvis}}}^{\left( i \right)},{\bm{p}_i}} \right] = S_{PN}^i\left( {{\bm{X^{\left( i \right)}}}} \right)
\end{equation}
where $S_{PN}^i\left(  \cdot  \right)$ is the $i$-th pose estimation module, $\bm{p_1} = \left[ {{\bm{p}_{{\text{Spine1}}}},{\bm{p}_{{\text{Spine2}}}},{\bm{p}_{{\text{Spine3}}}}} \right]$, ${\bm{p}_2} = \left[ {{\bm{p}_{\text{1}}},{\bm{p}_{{\text{d2}}}}} \right]$, ${\bm{p}_3} = \left[ {{\bm{p}_{\text{1}}},{\bm{p}_{{\text{d2}}}},{\bm{p}_{{\text{d3}}}}} \right]$.
Consequently, we derive the loss function ${{L}_{i}}$ as follows:

\begin{equation}
	{L_i} = \lambda \left\| \bm{\tilde p}_{{\text{pelvis}}}^{(i)} - \bm{p}_{{\text{pelvis}}}^{(i) \text{GT}} \right\|_2^2 + \left\| \bm{\tilde p}_i - \bm{p}_i^{GT} \right\|_2^2
\end{equation}
where $\bm{\tilde{p}}$ represents the estimated joint poses, and $\bm{p}^{GT}$ denotes the ground truth.

For the fourth stage, the output is the local rotation of the lower-body joints relative to the parent joints, expressed as ${\bm{p}_4} = S_{PN}^4\left( {{\bm{X}^{\left( 4 \right)}}} \right)$, where ${\bm{p}_{4}}={\bm{p}_{d4}}$. The loss function is expressed as:
\begin{equation}
	L_4 = \left\| \bm{\tilde p}_4 - \bm{p}_4^{GT} \right\|_2^2
\end{equation}

In addition, the pose is constrained not only by the relative rotation between joints but also by the positional relationships. Therefore, we utilize the estimated pose parameter to the skeletal hierarchy based on the SMPL model, and calculate the sub-joints global position ${\bm{b}_i}$ through Forward Kinematics, represented as:
\begin{equation}
	{\bm{b}_i} = FK\left( {\left[ {{\bm{p}_{{\text{pelvis}}}},{\bm{p}_i}} \right]} \right)
\end{equation}
where $FK(\cdot )$ is a forward kinematics function, which takes local joint rotation as input and outputs the position of the joint in the global coordinate system.

The integration of joint position information is more consistent with the fundamental biomechanical constraints compared to minimizing joint rotation alone. We simultaneously consider both rotation errors and position errors in backpropagation optimization to obtain the loss, expressed as:
\begin{equation}
	L_i = \lambda \left\| \bm{{\tilde p}}_{{\text{pelvis}}} - \bm{p}_{{\text{pelvis}}}^{\text{GT}} \right\|_2^2 + \left\| \bm{{\tilde p}}_i - \bm{p}_i^{\text{GT}} \right\|_2^2 + L_b
\end{equation}
\begin{equation}
	L_4 = \left\| \bm{{\tilde p}}_4 - \bm{p}_4^{\text{GT}} \right\|_2^2 + L_b
\end{equation}
where ${L_b} = \left\| {\bm{\tilde b} - {\bm{b}^{GT}}} \right\|_2^2$ represents joint position consistency loss. Referring to \cite{ref12}, the weight parameter $\lambda $ is set to 0.1 to balance the error of pelvic rotation (global orientation) with the error of local rotations of other joints, thereby ensuring the stability and convergence speed of the optimization process. The proposed integrated optimization method helps to alleviate the potential instability that may arise from relying solely on joint rotations, thereby improving the accuracy and biological plausibility of full-body pose estimation to achieve a more natural and realistic avatar representation.

\section{Experience}
\subsection{Dataset}
We use three public datasets widely recognized in the motion capture field for training, validation, and testing: AMASS \cite{ref31}, TotalCapture \cite{ref32}, and DIP-IMU \cite{ref9} datasets, as detailed in Table \ref{tab2}. Since the data size of TotalCapture and DIP-IMU datasets is insufficient for network training, we add additional synthetic inertial data generated by the AMASS dataset as a supplement to increase the diversity and quantity of training data \cite{ref10,ref24}. Following the same protocol as \cite{ref10}, we use the data collected by the last two participants in the DIP-IMU dataset for verification and the rest for training. In addition, HumanEval \cite{ref33} and Transition \cite{ref31} subsets in the AMASS dataset are used for testing, and the remaining subsets are used for training. Since the TotalCapture dataset is relatively limited in size and type, we use it only for testing as a check for cross-dataset generalization. Incorporating the DIP-IMU dataset containing real inertial measurements (with noise and drift) in the synthetic training data to fine-tune the network, thereby reducing the distribution discrepancies between synthetic and real data, which helps generalize to real-world application scenarios.
In addition, we recalibrate the acceleration measurements in TotalCapture to align the average acceleration measurement of each sequence with the average synthetic value. We also align the orientation of the AMASS dataset with the DIP-IMU dataset in the global frame to unify the character orientation.
\begin{table}[!t]
	\caption{Details of the AMASS, TotalCapture, and DIP-IMU datasets.}
	\label{tab2}
	\centering
	\setlength{\tabcolsep}{2mm}{
		\begin{tabular}{c|ccccc}
			\toprule
			Dataset & Type     & People & Motions & Frames & Minutes\\
			\midrule
			AMASS    & Synthetic & 487   & 14208 & 8122942 & 1335383  \\
			DIP-IMU & Real & 10   & 5 & 176198 & 2937   \\
			TotalCapture  &Real & 5   & 4 & 316779 & 5280     \\
			\bottomrule
	\end{tabular}}
\end{table}
\begin{table*}[!b]
	\centering
	\caption{Estimation and evaluation of full-body joint poses and comparison of online performance between ProgIP and baselines on the AMASS-HumanEval\&Transition and TotalCapture datasets.}
	\label{tab3}
	\setlength{\tabcolsep}{4mm}{
		\begin{tabular}{c|c|cccccc}
			\toprule
			Dataset & Method      & RE                & RE-Pelvis         & PE               & PE-Wrist         & Me               \\
			\midrule
			& IMUposer    & 17.40 (+/- 9.20)  & 15.21 (+/- 7.98)  & 9.45 (+/- 5.47)  & 14.48 (+/- 7.11) & 10.30 (+/- 5.81) \\
			& AGRoL       & 14.47 (+/- 7.79)  & 16.77 (+/- 8.13)  & 9.52 (+/- 5.48)  & 12.89 (+/- 5.76) & 10.05 (+/- 5.49) \\
			& TransPose   & 14.16 (+/- 7.89)  & 15.05 (+/- 7.98)  & 8.71 (+/- 5.74)  & 10.51 (+/- 6.46) & 8.86 (+/- 5.87)  \\
			& AvatarPoser & 12.65 (+/- 7.31)  & 14.01 (+/- 6.97)  & 7.49 (+/- 4.89)  & 8.48 (+/- 5.37)  & 7.55 (+/- 4.96)  \\
			\multirow{-5}{*}{AMASS} &
			\cellcolor[HTML]{EFEFEF}\textbf{ProgIP} &
			\cellcolor[HTML]{EFEFEF}\textbf{11.42 (+/- 6.35)} &
			\cellcolor[HTML]{EFEFEF}\textbf{13.79 (+/- 7.18)} &
			\cellcolor[HTML]{EFEFEF}\textbf{7.06 (+/- 4.88)} &
			\cellcolor[HTML]{EFEFEF}\textbf{7.87 (+/- 4.79)} &
			\cellcolor[HTML]{EFEFEF}\textbf{7.02 (+/- 4.74)} \\
			\midrule
			& IMUposer    & 19.44 (+/- 11.78) & 18.05 (+/- 12.03) & 11.34 (+/- 7.81) & 14.61 (+/- 8.77) & 12.23 (+/- 8.08) \\
			& AGRoL       & 19.18 (+/- 11.60) & 16.21 (+/- 10.98) & 10.01 (+/- 7.13) & 14.50 (+/- 8.74) & 10.98 (+/- 7.52) \\
			& TransPose   & 18.04 (+/- 11.13) & 16.31 (+/- 11.25) & 9.82 (+/- 7.25)  & 12.32 (+/- 7.88) & 10.27 (+/- 7.37) \\
			& AvatarPoser & 16.74 (+/- 10.53) & 14.83 (+/- 10.11) & 8.53 (+/- 6.47)  & 10.69 (+/- 6.80) & 8.97 (+/- 6.50)  \\
			\multirow{-5}{*}{TotalCapture} &
			\cellcolor[HTML]{EFEFEF}\textbf{ProgIP} &
			\cellcolor[HTML]{EFEFEF}\textbf{16.17 (+/- 9.98)} &
			\cellcolor[HTML]{EFEFEF}\textbf{14.01 (+/- 9.81)} &
			\cellcolor[HTML]{EFEFEF}\textbf{8.07 (+/- 6.22)} &
			\cellcolor[HTML]{EFEFEF}\textbf{10.33 (+/- 6.43)} &
			\cellcolor[HTML]{EFEFEF}\textbf{8.50 (+/- 6.18)} \\
			\bottomrule
	\end{tabular}}
\end{table*}

\subsection{Training Strategy}
During the training process, we feed the sequence with the input block size of $M$ frames to the network, and propagate the error only for the specific $N$-th frame back. The selected frame serves as the current frame in the real-time testing, meaning that the network uses the past $N-1$ frames and the future $M-N$ frames in the input sequence to estimate the current $N$-th frame. This strategy helps to improve the interpretability of the model, while reducing overfitting and saving computational resources especially when dealing with large-scale data.

\subsection{Implementation Details}
The training, evaluation, and testing of ProgIP are conducted using the PyTorch framework on a computer equipped with an AMD ${\text{Ryze}}{{\text{n}}^{{\text{TM}}}}$ 7 5700X CPU and an NVIDIA GeForce RTX 4060 Ti graphics card. We set the input block size to $M = 40$ and the current frame to $N = 30$, resulting in a tolerable latency of 166 milliseconds in the live demonstration. To ensure the full reproducibility of the network and the validity of the ablation experiments, the random seed was set to 10. We used the Adam optimizer \cite{ref34} with a batch size of 256 and a learning rate of ${10^{ - 4}}$ to optimize the network parameters. We employed the Noitom PN Lab system with three IMU sensors to collect real data, and the front end of the live demonstration was implemented in Unity3D. For subsequent practical applications, we standardized the frame rate to 60 Hz. Please note that we do not perform temporal filtering on the input data.

\subsection{Evaluation Metrics}
We quantitatively evaluated the proposed method using well-established metrics introduced in related work:
(1) Mean Joint Rotation Error [deg] (MJRE): The mean angular error of all joints between the estimated global rotation and the ground truth. MJRE-Pelvis evaluates the global rotation error of the pelvis.
(2) Mean Joint Position Error [cm] (MJPE): The mean Euclidean distance error of all joints between the estimated Cartesian positions and the ground truth, with the pelvic joint aligned. MJPE-Wrist evaluates the mean Euclidean distance error of both wrists.
(3) Mesh Error [cm] (ME): The mean Euclidean distance error of all mesh vertices of the SMPL model, with the pelvic joint aligned.

\subsection{Comparison with existing methods}
\begin{table*}[!t]
	\centering
	\caption{Estimation of full-body joint poses and evaluation of upper-body joint poses, and comparison of online performance between ProgIP and baselines on the AMASS-HumanEval\&Transition and TotalCapture datasets.}
	\label{tab4}
	\setlength{\tabcolsep}{4mm}{
		\begin{tabular}{c|c|cccccc}
			\toprule
			Dataset & Method      & RE                & RE-Pelvis         & PE               & PE-Wrist         & Me               \\
			\midrule
			& IMUposer    & 15.31 (+/- 7.83)  & 15.21 (+/- 7.98)  & 8.69 (+/- 4.68)  & 14.48 (+/- 7.11) & 9.85 (+/- 5.33)  \\
			& AGRoL       & 12.82 (+/- 6.29)  & 16.77 (+/- 8.13)  & 9.04 (+/- 4.67)  & 12.89 (+/- 5.76) & 9.78 (+/- 5.00)  \\
			& TransPose   & 12.00 (+/- 6.44)  & 15.05 (+/- 7.98)  & 8.02 (+/- 5.12)  & 10.51 (+/- 6.46) & 8.46 (+/- 5.50)  \\
			& AvatarPoser & 10.93 (+/- 5.89)  & 14.01 (+/- 6.97)  & 7.04 (+/- 4.13)  & 8.48 (+/- 5.37)  & 7.29 (+/- 4.49)  \\
			\multirow{-5}{*}{AMASS} &
			\cellcolor[HTML]{EFEFEF}\textbf{ProgIP} &
			\cellcolor[HTML]{EFEFEF}\textbf{9.78 (+/- 4.92)} &
			\cellcolor[HTML]{EFEFEF}\textbf{13.79 (+/- 7.18)} &
			\cellcolor[HTML]{EFEFEF}\textbf{6.87 (+/- 4.00)} &
			\cellcolor[HTML]{EFEFEF}\textbf{7.87 (+/- 4.79)} &
			\cellcolor[HTML]{EFEFEF}\textbf{6.92 (+/- 4.19)} \\
			\midrule
			& IMUposer    & 17.28 (+/- 10.44) & 18.05 (+/- 12.03) & 10.59 (+/- 6.90) & 14.61 (+/- 8.77) & 11.78 (+/- 7.52) \\
			& AGRoL       & 16.97 (+/- 10.25) & 16.21 (+/- 10.98) & 8.89 (+/- 6.07)  & 14.50 (+/- 8.74) & 10.28 (+/- 6.86) \\
			& TransPose   & 15.68 (+/- 9.62)  & 16.31 (+/- 11.25) & 8.66 (+/- 6.15)  & 12.32 (+/- 7.88) & 9.55 (+/- 6.68)  \\
			& AvatarPoser & 14.72 (+/- 8.94)  & 14.83 (+/- 10.11) & 7.69 (+/- 5.30)  & 10.69 (+/- 6.80) & 8.45 (+/- 5.76)  \\
			\multirow{-5}{*}{TotalCapture} &
			\cellcolor[HTML]{EFEFEF}\textbf{ProgIP} &
			\cellcolor[HTML]{EFEFEF}\textbf{14.17 (+/- 8.40)} &
			\cellcolor[HTML]{EFEFEF}\textbf{14.01 (+/- 9.81)} &
			\cellcolor[HTML]{EFEFEF}\textbf{7.27 (+/- 5.10)} &
			\cellcolor[HTML]{EFEFEF}\textbf{10.33 (+/- 6.43)} &
			\cellcolor[HTML]{EFEFEF}\textbf{8.00 (+/- 5.47)} \\
			\bottomrule
	\end{tabular}}
\end{table*}
\begin{table*}[!t]
	\centering
	\caption{Estimation and evaluation of upper-body joint poses and comparison of online performance between ProgIP and baselines on the AMASS-HumanEval\&Transition and TotalCapture datasets.}
	\label{tab5}
	\setlength{\tabcolsep}{4mm}{
		\begin{tabular}{c|c|cccccc}
			\toprule
			Dataset & Method      & RE                & RE-Pelvis         & PE              & PE-Wrist         & Me               \\
			\midrule
			& IMUposer    & 13.87 (+/- 7.65)  & 14.37 (+/- 8.23)  & 7.92 (+/- 4.70) & 12.67 (+/- 6.90) & 8.87 (+/- 5.30)  \\
			& AGRoL       & 11.77 (+/- 5.92)  & 15.61 (+/- 7.99)  & 8.45 (+/- 4.54) & 11.66 (+/- 5.49) & 9.08 (+/- 4.86)  \\
			& TransPose   & 12.20 (+/- 6.26)  & 15.52 (+/- 7.83)  & 8.17 (+/- 5.01) & 10.49 (+/- 6.23) & 8.55 (+/- 5.38)  \\
			& AvatarPoser & 11.12 (+/- 5.94)  & 13.98 (+/- 7.07)  & 7.08 (+/- 4.05) & 9.01 (+/- 5.50)  & 7.40 (+/- 4.42)  \\
			\multirow{-5}{*}{AMASS} &
			\cellcolor[HTML]{EFEFEF}\textbf{ProgIP} &
			\cellcolor[HTML]{EFEFEF}\textbf{9.78 (+/- 4.92)} &
			\cellcolor[HTML]{EFEFEF}\textbf{13.79 (+/- 7.18)} &
			\cellcolor[HTML]{EFEFEF}\textbf{6.87 (+/- 4.00)} &
			\cellcolor[HTML]{EFEFEF}\textbf{7.87 (+/- 4.79)} &
			\cellcolor[HTML]{EFEFEF}\textbf{6.92 (+/- 4.19)} \\
			\midrule
			& IMUposer    & 16.66 (+/- 10.18) & 17.10 (+/- 11.85) & 9.86 (+/- 6.71) & 13.77 (+/- 8.44) & 11.03 (+/- 7.28) \\
			& AGRoL       & 16.14 (+/- 9.59)  & 15.57 (+/- 10.29) & 8.42 (+/- 5.75) & 13.35 (+/- 8.22) & 9.66 (+/- 6.48)  \\
			& TransPose   & 15.72 (+/- 9.45)  & 16.32 (+/- 11.13) & 8.58 (+/- 6.06) & 12.07 (+/- 7.87) & 9.44 (+/- 6.58)  \\
			& AvatarPoser & 14.90 (+/- 9.09)  & 14.97 (+/- 10.33) & 7.70 (+/- 5.39) & 10.98 (+/- 7.09) & 8.48 (+/- 5.85)  \\
			\multirow{-5}{*}{TotalCapture} &
			\cellcolor[HTML]{EFEFEF}\textbf{ProgIP} &
			\cellcolor[HTML]{EFEFEF}\textbf{14.17 (+/- 8.40)} &
			\cellcolor[HTML]{EFEFEF}\textbf{14.01 (+/- 9.81)} &
			\cellcolor[HTML]{EFEFEF}\textbf{7.27 (+/- 5.10)} &
			\cellcolor[HTML]{EFEFEF}\textbf{10.33 (+/- 6.43)} &
			\cellcolor[HTML]{EFEFEF}\textbf{8.00 (+/- 5.47)} \\
			\bottomrule
	\end{tabular}}
\end{table*}
\begin{figure*}[!b]
	\centering
	\includegraphics[width=7.1in]{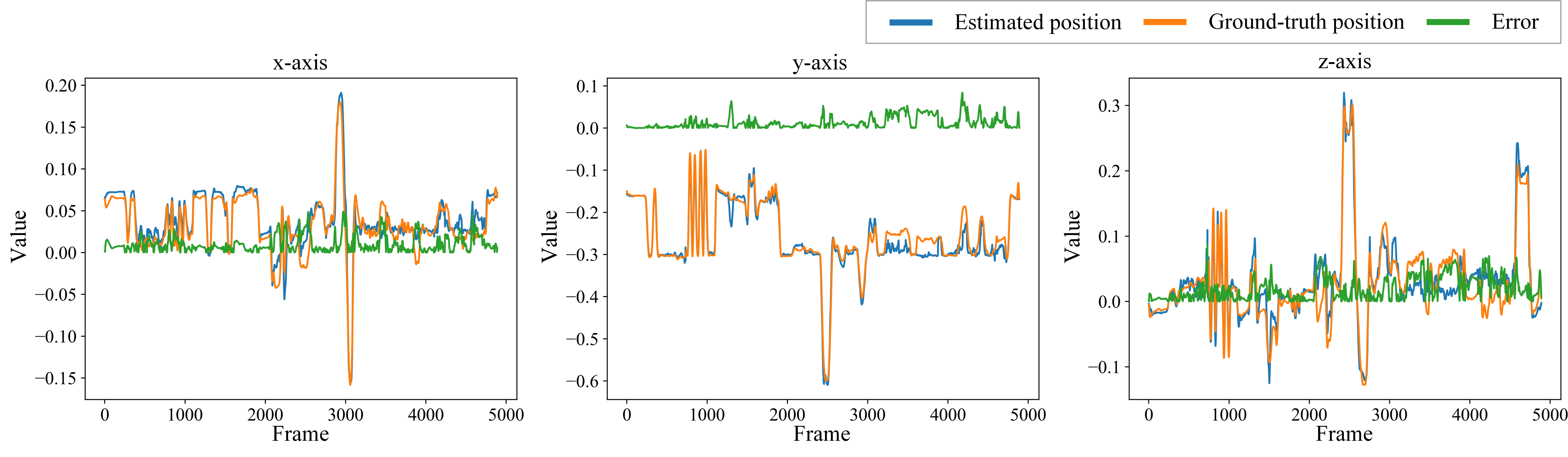}
	\caption{The mean position error of the full-body joints along the x-axis, y-axis, and z-axis of the partial sequence in the TotalCapture dataset. The blue line represents the mean estimated joint position, the orange line represents the mean ground truth joint position, and the green line represents the average position error.}
	\label{fig5}
\end{figure*}
We selected four baselines that are most similar to our work from recent state-of-the-art methods for estimating full-body poses from sparse inputs. The first baseline is AvatarPoser. Since our input does not include positional data, we adjust the input signals to include acceleration, rotation, and angular velocity, while ignoring the inverse kinematics module. The second baseline is AGRoL, for which we also adjust its input to acceleration, rotation, and angular velocity. IMUPoser is closest to our method due to its perfect match with the device combinations mentioned, and we omit the downsampling and filtering of the input signal. The final baseline is TransPose, which uses six IMU sensors worn at specific locations. Therefore, we remove the sensors worn on the pelvis and lower-body, estimating only the upper-body joint positions as an intermediate process, without considering global translation. All baselines are publicly available on GitHub. For a fair comparison, we follow the original implementation for training, validation, and testing on the same datasets, and maintain other details consistent with the original papers.

\subsubsection{Quantitative evaluation}
To demonstrate the effectiveness of the proposed ProgIP, we quantitatively compare it with four baselines using test sequences from existing datasets (AMASS-HumanEval\&Transition and TotalCapture). Considering that the quality of upper-body representation is also crucial for virtual reality applications, we divide the quantitative evaluation into three scenarios: estimating and evaluating full-body joint poses, estimating full-body joint poses but evaluating only upper-body joint poses, and estimating and evaluating upper-body joint poses. These results are detailed in Table \ref{tab3}, Table \ref{tab4} and Table \ref{tab5}, respectively. We report the mean and standard deviation for each metric, with ProgIP achieving the best results across all metrics and outperforming the four baselines.
\begin{figure}[!tb]
	\centering
	\includegraphics[width=3.4in]{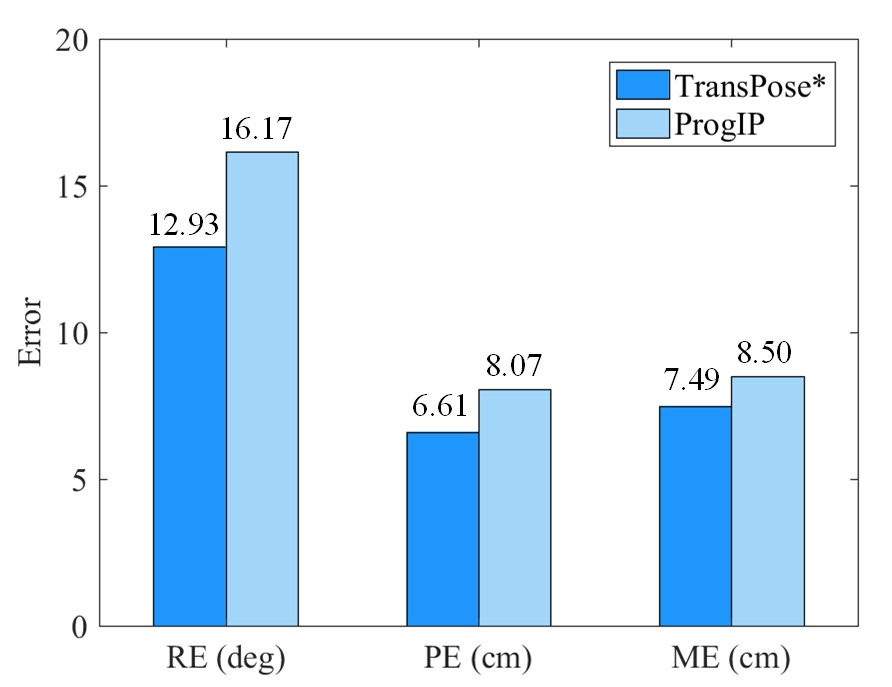}
	\caption{Performance comparison of ProgIP and the original TransPose using six IMU sensors. The superscript * denotes the original paper.}
	\label{fig6}
\end{figure}
\begin{figure*}[!b]
	\centering
	\includegraphics[width=7.15in]{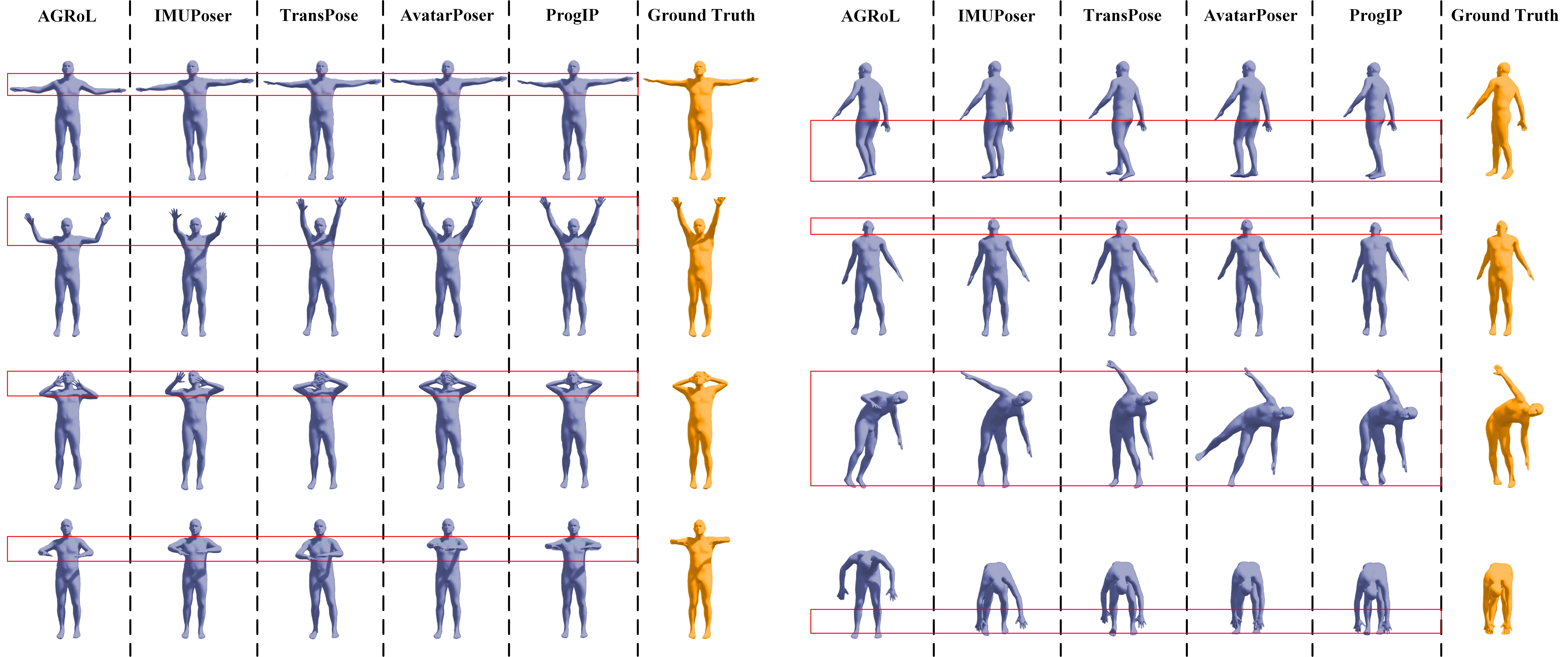}
	\caption{Qualitative comparison of our method ProgIP with four baselines. We conduct online comparison on the TotalCapture dataset and select some results here, where the orange is ground truth. Additional qualitative results are available in the supplementary materials.}
	\label{fig7}
\end{figure*}
AvatarPoser is inferior to our method and achieves the second-best performance on both datasets, where the Transformer-based network provides a significant advantage and the forward kinematics module reduces the accumulation of rotation errors in the kinematic chain. However, AvatarPoser directly estimates full-body pose from input signals and relies on a single Transformer architecture to extract global features, without explicitly modeling the hierarchical relationships of joints. The third place is TransPose, which uses the joint position as the intermediate process to solve the relative rotation of the joints. However, relying solely on three sets of inertial measurements is insufficient for accurately estimating the root relative position of the joint. IMUPoser achieved the second-to-last result in TotalCapture and the worst result in AMASS. Compared with TransPose, it simplifies the solution of joint positions and the designed RNN structure is relatively simple. AGRoL performs the worst in all metrics in TotalCapture and the second-to-last performance in AMASS, due to its MLP-based diffusion model. Although the specially customized motion-conditioned diffusion model plays a key role in motion generation, its MLP backbone does not adequately capture temporal information. 
Fig. \ref{fig5} shows the mean position errors of the full-body joints along the x-axis, y-axis, and z-axis for the partial sequences in the TotalCapture dataset. It can be seen that the joint error does not drift significantly over time, but is only related to the action of the current frame. This is attributed to multi-stage progressive estimation and joint position consistency loss designed by ProgIP, which enhances the dependency between adjacent joints and reduces the accumulation of joint rotation estimation errors along the kinematic chain.
When tested on the TotalCapture real dataset, ProgIP performs similarly to the original TransPose, with rotation error differing by 3.24 deg, global position error by 1.46 cm, and mesh position error by 1.01 cm, which is close to the full-body pose estimation scheme using six IMU sensors, as shown in Fig. \ref{fig6}.

Additionally, the error margins of ProgIP for different types of motion are specifically reported to demonstrate its reliability. We conduct experiments on the TotalCapture dataset, including three replicates for each of the four motion types, and report the performance and error margins for different motion types, as shown in Table \ref{tab11}.
\begin{table}[htb]
	\caption{Error margins for different motion types.}
	\label{tab11}
	\setlength{\tabcolsep}{1.4mm}{
		\begin{tabular}{c|ccccc}
			\toprule
			Motion & RE & RE-Pelvis & PE & PE-Wrist & ME \\
			\midrule
			Acting1& 15.4632& 12.2639&7.1213&9.9872&7.5915\\
			Acting2&16.3864&12.0632&7.1879&9.7243&7.6254\\
			Acting3&16.8083&14.5456&8.4185&10.4747&8.8982\\
			Freestyle1&18.8902&19.6545&11.1892&12.7348&	11.4093\\
			Freestyle2&21.0383&19.4369&11.7885&13.7456&	12.1566\\
			Freestyle3&28.7018&24.0060&14.7723&	22.1103&	16.3339\\
			Rom1&13.1808&10.5633&6.2621&6.8497&6.3128\\
			Rom2&12.6847&9.6734&5.6447&6.5468&5.8281\\
			Rom3&12.2427&10.8259&5.6481&6.4169&5.7638\\
			Walking1&10.4948&8.2597&4.8203&6.5451&5.0911\\
			Walking2&11.5415&10.5645&5.2881&6.8734&5.6125\\
			Walking3&11.9107&10.1390&5.5036&7.2858&5.8703\\
			\midrule
			Mean&15.7786&13.4997&7.8037&9.9412&8.2078\\
			Standard deviation&5.1834&4.9173&3.1526&4.5944&3.4406\\
			\bottomrule
	\end{tabular}}
\end{table}

\subsubsection{Qualitative evaluation}
We use partial sequences selected from the TotalCapture dataset to compare the poses reconstructed by ProgIP with those of the four baselines, and the qualitative results from the real dataset better reflect the stability and superiority of ProgIP. Fig. \ref{fig7} intuitively presents some examples where ProgIP demonstrates superior performance and effectively captures the nuances of challenging motions, especially arm motions and pelvic rotation. However, for lower-body reconstruction of turning motions, our results remain reasonable even when the estimated leg poses slightly differ from the ground truth. In some scenarios, we see that ProgIP successfully reconstructs both the upper and lower body, while AGRoL fails to accurately estimate upper arm poses in certain cases. The performance of ProgIP with these real data can be attributed to the well-designed encoder and decoder help capture both consistency and variation in motion, combined with progressive body modeling, which is particularly beneficial for estimating challenging poses. As evidenced by the qualitative results, we achieve visually pleasing state-of-the-art online capture quality.

A substantial number of quantitative and qualitative experimental results demonstrate that ProgIP significantly outperforms baselines in terms of capture accuracy and physical realism. In the progressive estimation of depth along the kinematic chain, the TE-biLSTM encoder and the MLP-based decoder are used to better capture state change signals to resolve motion blur. At the same time, the further improvement in estimation accuracy is attributed to the effective constraint of joint positions calculated using forward kinematics.

\subsection{Ablation experiment}
\begin{table*}[!t]
	\centering
	\caption{Ablation studies on body region division, progressive estimation, global feature extraction module and forward kinematics. It shows the contribution of our key components to pose estimation.}
	\label{tab6}
	\setlength{\tabcolsep}{3.5mm}{
		\begin{tabular}{c|ccc|ccc}
			\toprule
			& \multicolumn{3}{c|}{AMASS}                                    & \multicolumn{3}{c}{TotalCapture}                                         \\ \cmidrule(l){2-7} 
			\multirow{-2}{*}{Method} & \multicolumn{1}{c}{RE} & PE                & ME              & \multicolumn{1}{c}{RE} & \multicolumn{1}{c}{PE} & \multicolumn{1}{c}{ME} \\
			\midrule
			No deep-based region                     & 12.19 (+/- 6.60)       & 7.42 (+/- 4.85)   & 7.38 (+/- 4.72) & 16.62 (+/- 10.13)      & 8.24 (+/- 6.41)        & 8.65 (+/- 6.37)        \\
			No progres               & 12.86 (+/- 7.20)       & 7.47 (+/- 4.82) & 7.46 (+/- 4.77) & 16.61 (+/- 10.31)      & 8.45 (+/- 6.38)        & 8.89 (+/- 6.40)        \\
			No global                & 12.43 (+/- 6.82)       & 7.72 (+/- 5.14)   & 7.79 (+/- 5.09) & 17.30 (+/- 10.55)      & 8.93 (+/- 6.82)        & 9.34 (+/- 6.82)        \\
			No FK                    & 12.31 (+/- 6.57)       & 7.58 (+/- 4.88)   & 7.57 (+/- 4.79) & 16.97 (+/- 10.49)      & 8.75 (+/- 6.71)        & 9.18 (+/- 6.73)        \\
			\rowcolor[HTML]{EFEFEF} 
			\textbf{ProgIP} &
			\textbf{11.42 (+/- 6.35)} &
			\multicolumn{1}{c}{\cellcolor[HTML]{EFEFEF}\textbf{7.06 (+/- 4.88)}} &
			\multicolumn{1}{c|}{\cellcolor[HTML]{EFEFEF}\textbf{7.02 (+/- 4.74)}} &
			\textbf{16.17 (+/- 9.98)} &
			\textbf{8.07 (+/- 6.22)} &
			\textbf{8.50 (+/- 6.18)} \\
			\bottomrule 
	\end{tabular}}
\end{table*}
To evaluate the effectiveness of the key components of ProgIP, we compare it with four additional variants: (1) No deep-based region: the body is segmented into three regions using the body region segmentation technique used in \cite{ref24} without considering kinematic chain constraints; (2) No progress: the full-body poses are directly estimated using inertial measurements rather than the multi-stage progressive estimation; (3) No global: the progressive estimation task relies solely on inertial measurements without global information; (4) No FK: the loss function only minimizes the rotation angles without incorporating additional constraints from joint positions calculated by forward kinematics. We compare these four variants with our method on the AMASS-HumanEval\&Transition and Total Capture datasets, and the experimental results in Table \ref{tab6} clearly show the performance differences. The removal of these components significantly increases joint rotation and position errors. ProgIP progressively estimates descendant joint poses and iteratively updates parent joint poses in order to increase kinematic chain depth, which positively contributes to optimizing full-body motion reconstruction. Additionally, we constrain joint rotations relative to parent joints using positions calculated by forward kinematics to further improve performance. Trends in both datasets confirm that ProgIP not only performs well on synthetic data but is also robust and effective in handling complex and dynamic motions in real-world scenarios.

\subsection{Network structure comparison}
\begin{table}[!tb]
	\centering
	\caption{Performance comparison of our proposed encoder with the Transformer Encoder and biLSTM}
	\label{tab7}
	\setlength{\tabcolsep}{2.8mm}{
		\begin{tabular}{c|ccc}
			\toprule
			& \multicolumn{3}{c}{AMASS}  \\ \cmidrule(l){2-4} 
			\multirow{-2}{*}{Method} & \multicolumn{1}{c}{RE} & PE              & ME           \\
			\midrule
			TF                       & 12.19 (+/- 6.69)       & 7.68 (+/- 5.08) & 7.63 (+/- 4.97) \\
			biLSTM                   & 13.88 (+/- 7.81)       & 8.33 (+/- 5.56) & 8.38 (+/- 5.63) \\
			\rowcolor[HTML]{EFEFEF} 
			\textbf{ProgIP} &
			\textbf{11.42 (+/- 6.35)} &
			\textbf{7.06 (+/- 4.88)} &
			\textbf{7.02 (+/- 4.74)} \\
			\bottomrule
			\toprule
			& \multicolumn{3}{c}{TotalCapture}  \\ \cmidrule(l){2-4} 
			\multirow{-2}{*}{Method} & \multicolumn{1}{c}{RE} & PE              & ME           \\
			\midrule
			TF                       & 17.11 (+/- 10.41)      & 8.93 (+/- 6.78)        & 9.34 (+/- 6.74) \\
			biLSTM                   & 17.89 (+/- 11.31)      & 9.74 (+/- 7.19)        & 10.25 (+/- 7.34) \\
			\rowcolor[HTML]{EFEFEF} 
			\textbf{ProgIP} &
			\textbf{16.17 (+/- 9.98)} &
			\textbf{8.07 (+/- 6.22)} &
			\textbf{8.50 (+/- 6.18)} \\
			\bottomrule
	\end{tabular}}
\end{table}

The choice of network structure plays a vital role in encoder performance, so we compare the network components of the designed encoder to highlight the advantages of the TE-biLSTM encoder. This section considers two popular alternative backbone architectures: Transformer and RNN, and evaluates their performance on the pose estimation task. To ensure a fair comparison, the input of the alternative architecture used is consistent with the inertial measurement $\bm{X} = \in {\mathbb{R}^{S \times 15}}$, and the feature dimension is extended to 256 through the FC layer. RNN architecture processes the motion data through a two-layer biLSTM network with a width of 256, and the feature dimension of the concatenated output of bidirectional hidden states is 512. Transformer architecture outputs features of the same dimension through three layers of multi-head self-attention (the number of heads is set to 8). Finally, the corresponding decoder projects the output features into the target space respectively.
As shown in Table \ref{tab7}, the TE-biLSTM encoder in ProgIP achieves better results than the other two backbone architectures. The RNN network exhibits significant rotation and position errors at some joints. The rotation and position errors in the AMASS dataset are 10.64\% and 20.69\% higher than those of the TE-biLSTM encoder, respectively, while in the TotalCapture dataset, these errors are 21.54\% and 17.99\% higher than those of the TE-biLSTM encoder, respectively. In comparison, the Transformer architecture has 5.81\% and 6.74\% higher rotation errors, and 10.66\% and 9.40\% higher position errors than the TE-biLSTM encoder. Using either the RNN or Transformer architecture alone leads to performance degradation, this is due to the limitations of their single and constrained data processing approaches. RNN architecture primarily captures local motion continuity explicitly, while Transformer architecture focuses on global dependencies through self-attention mechanisms. The architecture based on the combination of Transformer and RNN can better capture the dynamic features in time series data and improve the accurate estimation of joint poses. This demonstrates that the designed TE-biLSTM encoder combined with Transformer and RNN architecture benefits the reconstruction accuracy of full-body motion.
\begin{figure*}[!t]
	\centering
	\includegraphics[width=7.15in]{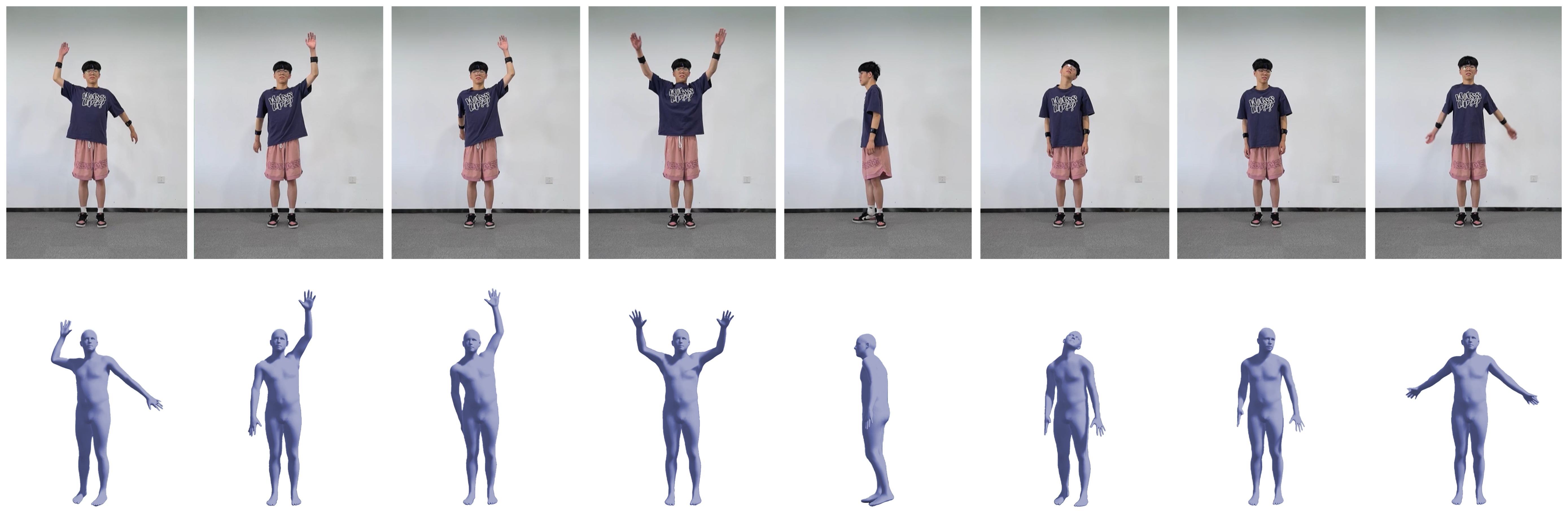}
	\caption{We test our method using motion data recorded by Noitom’s PN LAB inertial sensors and display the real-time animation rendering in Unity. The top shows the real-world motion performed by the user, and the bottom shows the rendered predicted animation. Please refer to our supplemental materials for more results.}
	\label{fig8}
\end{figure*}

\subsection{Live Demo}
We use the wireless high-precision pure inertial sensor system Perception Neuron Laboratory (PN Lab) developed by Noitom for live demonstrations. The sampling rate is set to 60Hz and the built-in AHRS calculation library provides sensor attitude data. The specific parameters shown in Table \ref{tab8}. Three Noitom sensors worn on the participants' heads and wrists are connected to a real-time data processing system on the computer via Bluetooth technology, and real-time animations are rendered in Unity. We conducted a series of real-time live demonstrations to validate the effectiveness and feasibility of ProgIP in full-body pose estimation. Specifically, a total of five independent experiments are conducted in the live demonstration, each lasting about two minutes and covering different motion scenarios. Since the similar reliability and consistency of each experiment are similar, the live demonstration shown uses the results of the first experiment. In the current version, one participant (male, 180cm tall, weighing 70kg) performs in the demonstration and tested various of full-body motions, including but not limited to walking, running, turning, and waving. These cover most of the basic motion patterns in daily life. The participant repeats each motions multiple times and the estimation results are very similar, which ensures the stability and repeatability of the live demonstration.

The system shows excellent stability during long-term operation, capable of generating smooth animation transitions in real-time without noticeable jitter or drift. The motions of the virtual characters are natural and realistic, in line with the laws of human kinematics, especially in terms of lower-body motions and pelvic rotation. Based on this, we generally believe that the virtual characters are close to real human bodies, and the overall performance is realistic and smooth.
\begin{table}[tb]
	\caption{Specific parameters of PN Lab inertial sensors.}
	\label{tab8}
	\setlength{\tabcolsep}{3.6mm}{
		\begin{tabular}{c|ccc}
			\toprule
			& Accelerometer & Gyroscope     & Magnetometer \\
			\midrule
			Range & ±8g           & ±2000 deg/s & ±10 Gauss    \\
			Accuracy & 0.244mg       & 0.07 deg/s  & 0.003 Gauss  \\
			\bottomrule
	\end{tabular}}
\end{table}
	
\subsection{Failure cases}
\begin{figure}[!htb]
	\centering
	\includegraphics[width=3.3in]{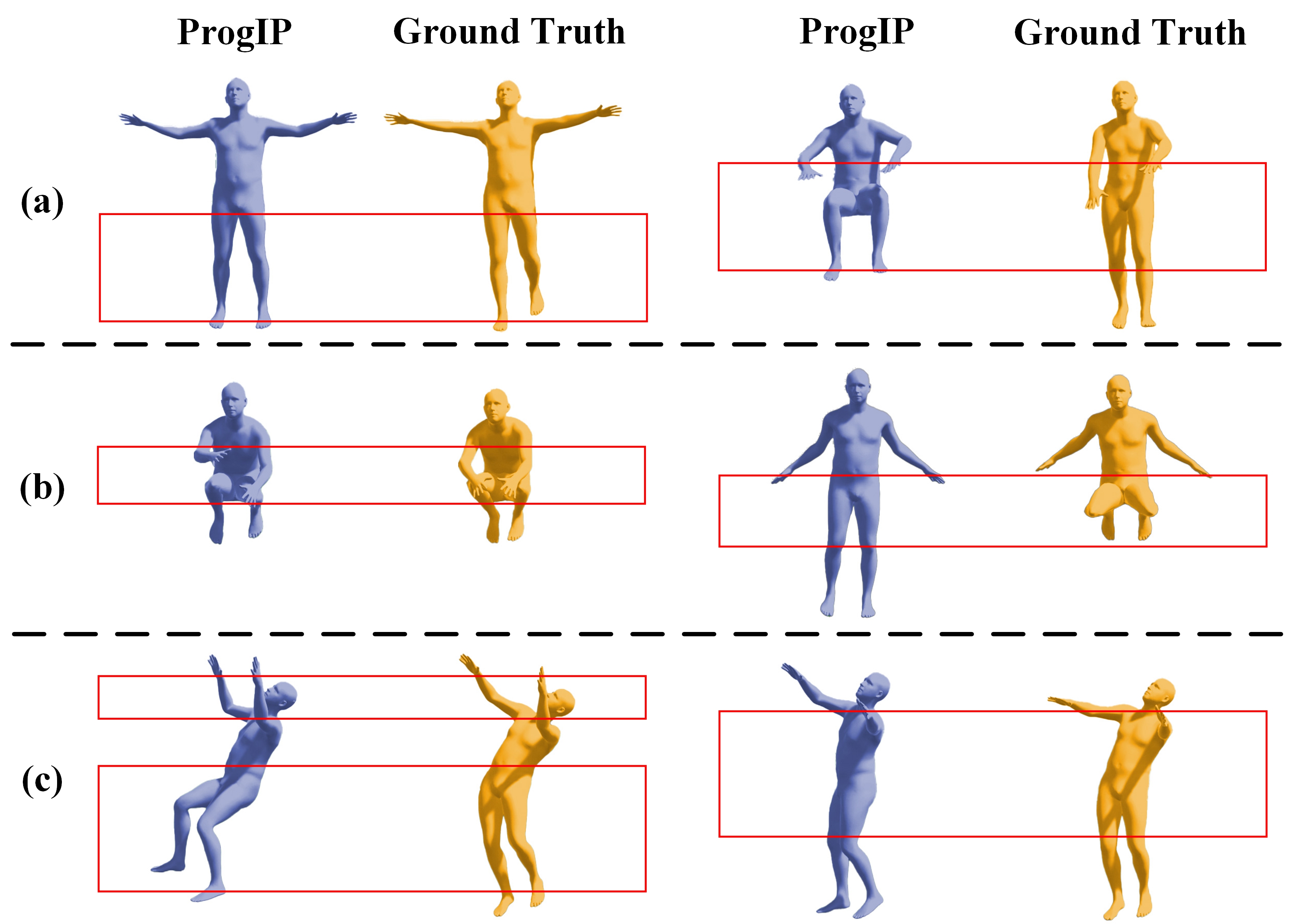}
	\caption{Failed cases. We show the comparative differences between the estimated pose and the ground-truth pose, including three specific motions: (a)Unconventional lower-body motions, (b)Sitting and standing up, (c)Fast and complex motions.}
	\label{fig11}
\end{figure}

We conduct a qualitative analysis of ProgIP on the TotalCapture dataset and find that it performed poorly in the following specific motions, as shown in the Fig. \ref{fig11}. We analyze it and clarified the direction of improvement, and subsequent work will focus on optimization:
(1) Unconventional lower-body motions: When the test motions are insufficiently covered in the training or the correlation between the motions of lower-body and upper-body is weak, the system may exhibit bias. For example, as shown in Fig. \ref{fig11}(a), swinging the arms up and down while backing up may lead to inaccurate leg prediction and even misidentification as jumping.
Enriching the training samples and introducing physical constraints on the feet may be a potential solution;
(2) Sitting and standing up: Since the rotation measurements of the pelvis and lower limbs are similar, it is difficult for the system to distinguish the details of the motion by relying only on the head and wrist IMU. For example, as shown in Fig. \ref{fig11}(b), when going up the stairs and squatting, the system estimates the correct pose for a short time, but then returns to the standing. Future work will explore a dynamic initial state encoder and an initial state consistency to improve the sensitivity to acceleration information;
(3) Fast and complex motions: When the subject suddenly changes his posture and moves drastically, the system may have a short-term posture abnormality, but it can gradually return to normal in a short time, as shown in Fig. \ref{fig11}(c). This may be due to insufficient diversity of training data and weak correlation of window data. We will consider fast or complex motions in training and improve the online strategy to improve robustness.

\subsection{Limitations and Future Work}
Firstly, ProgIP is a learning-based method, so the generated avatar animation may exhibit unnatural motions when encountering poses significantly different from the training datasets, such as shaking or foot sliding, but the poses generated by our method are nearly identical and reasonable. We will build and integrate representative and diverse datasets with real inertial data to enhance the generalization ability of the model in future work. Secondly, ProgIP may reconstruct inaccurate poses for motions such as sitting down and standing up, which have almost similar rotation measurements. Therefore, future work will explore an acceleration-based dynamic initial state encoder applied to the RNN architecture and introduce an initial state consistency regularization term in back propagation to further enhance the sensitivity to acceleration information. Thirdly, although ProgIP has a lower wrist position error compared to advanced baselines, there are still noticeable discrepancies from the ground truth in some cases. In the future, an effective compensation mechanism should be developed to optimize hand position estimation, because the hand position is crucial in virtual reality applications. Finally, pose estimation methods usually need to be applied across various practical scenarios and environments. Thus, integrating pose estimation technology with specific application contexts and addressing practical needs is an important issue to consider.

\section{Conclusions}
This paper introduces ProgIP, a pose estimation method that combines a human dynamics model with neural networks and uses only three IMU sensors worn on the head and wrists. ProgIP progressively reconstructs full-body motion by increasing the kinematic chain depth, with the TE-biLSTM encoder and MLP-based decoder effectively learning and mapping the temporal correlation features of human motion. Extensive experiments on multiple public datasets demonstrate that ProgIP outperforms advanced methods and meets the requirements for real-time operation by generating realistic and plausible motions. The proposed solution relying only on three IMU sensors provides economical and stable technical support for practical full-body virtual reality applications.

\section*{Acknowledgments}
This work was supported by the Fundamental Research Funds for the Provincial Universities of Zhejiang (GK259909299001-023), the Key R\&D Program of Zhejiang under Grant (2025C03001, 2023C01044), the National Nature Science Foundation of China (62301198), the National Key R\&D Program of China (2023YFC3305600), the Zhejiang Provincial Natural Science Foundation (LDT23F02024F02), and the NSFC (61822111, 62021002). This work was also supported by THUIBCS, Tsinghua University, and BLBCI, Beijing Municipal Education Commission.

\newpage
\begin{IEEEbiography}[{\includegraphics[width=1in,height=1.25in,clip,keepaspectratio]{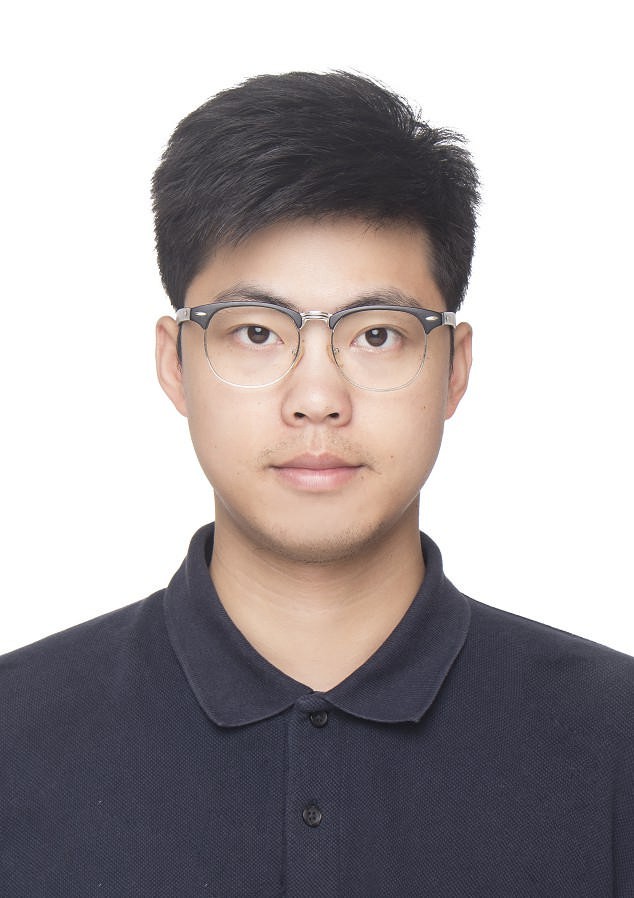}}]{Zunjie Zhu}
received the B.S. degree in electronic and information engineering and the Ph.D. degree in automation from Hangzhou Dianzi University, Hangzhou, China, in 2016 and 2022, respectively. He is currently an Assistant Professor with the Department of Communication Engineering, Hangzhou Dianzi University. His research interests include 3-D vision, simultaneous localization and mapping (SLAM), and image restoration.
\end{IEEEbiography}

\begin{IEEEbiography}[{\includegraphics[width=1in,height=1.25in,clip,keepaspectratio]{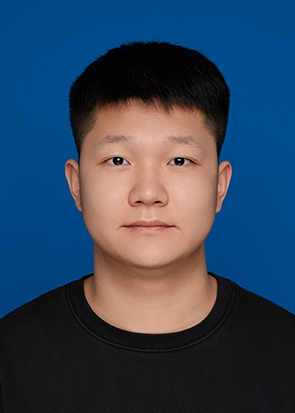}}]{Yan Zhao}
received the B.S. degree in electronic information engineering and the M.S. degree in electronic information from Linyi University, Linyi, China, in 2019 and 2023, respectively. He is currently pursuing the Ph.D. degree with the School of Control Science and Engineering, Tiangong University, Tianjin, China. His current research interests include pattern recognition, body sensor network, and intelligent signal processing.
\end{IEEEbiography}

\begin{IEEEbiography}[{\includegraphics[width=1in,height=1.25in,clip,keepaspectratio]{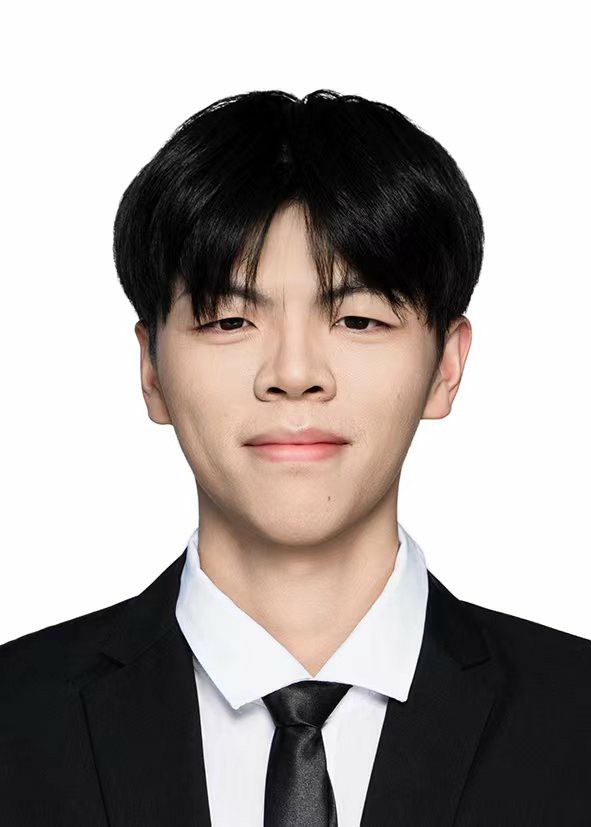}}]{Yihan Hu}
received the B.S. degree in communication engineering from Hangzhou Dianzi University, Hangzhou, China, in 2024. He is currently pursuing the master's degree with the College of Communication Engineering, Hangzhou Dianzi University, Hangzhou. His research interests include 3-D vision, simultaneous localization and mapping (SLAM), and image restoration.
\end{IEEEbiography}

\begin{IEEEbiography}[{\includegraphics[width=1in,height=1.25in,clip,keepaspectratio]{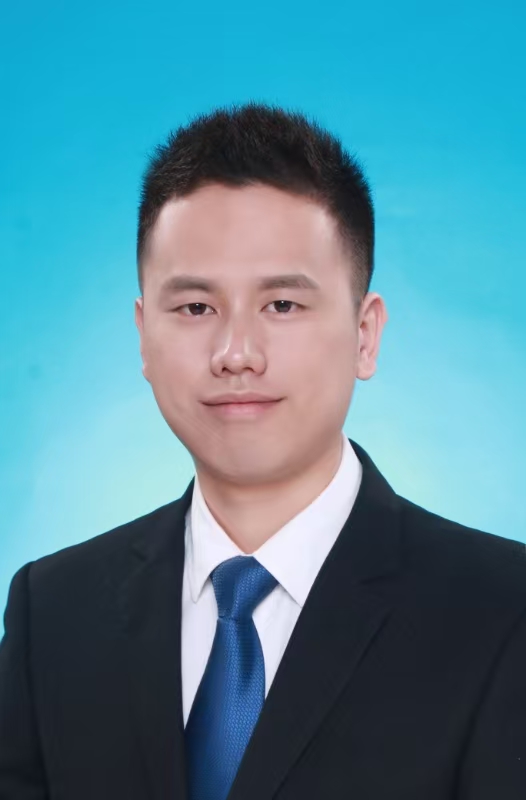}}]{Guoxiang Wang}
received the M.S. degree in Beijing University of Posts and Telecommunications in 2017. He is currently an Assistant Professor with Lishui University. His research interests are computer vision, pattern recognition, image processing.
\end{IEEEbiography}

\begin{IEEEbiography}[{\includegraphics[width=1in,height=1.25in,clip,keepaspectratio]{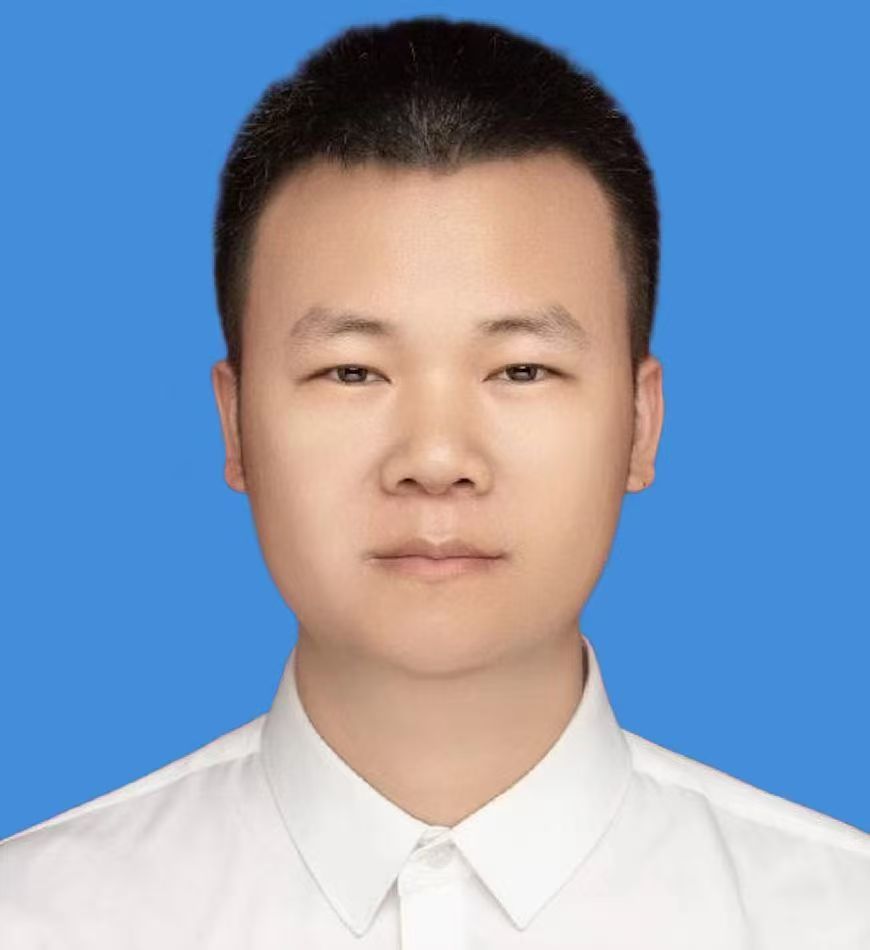}}]{Hai Qiu}
received B.S degree in industrial engineering from Southwest Jiaotong University, China, in 2009, M.S degree in industrial engineering from Shanghai Jiaotong University, China, in 2012, and PhD degree in human and system engineering from Ulsan National Institute of Science and Technology, South Korea, in 2016. He is currently an Senior Engineer with the Department of Visual Intelligence in Costar (Hang Zhou) Intelligent Optoelectronics  Technology Co.,Ltd. His research interests include machine learning, deep learning, and simultaneous localization and mapping (SLAM).
\end{IEEEbiography}

\begin{IEEEbiography}[{\includegraphics[width=1in,height=1.25in,clip,keepaspectratio]{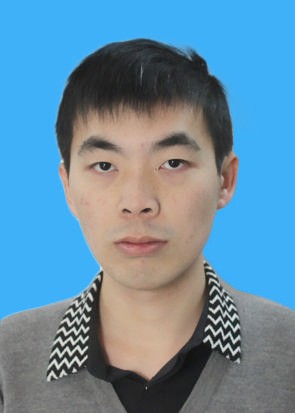}}]{Bolun Zheng}
received the B.S. and Ph.D. degrees in electronic information technology and instrument from Zhejiang University in 2014 and 2019, respectively. He is currently an Associate Professor with Hangzhou Dianzi University. His research interests are computer vision, pattern recognition, image processing and embedded parallel computing.
\end{IEEEbiography}

\begin{IEEEbiography}[{\includegraphics[width=1in,height=1.25in,clip,keepaspectratio]{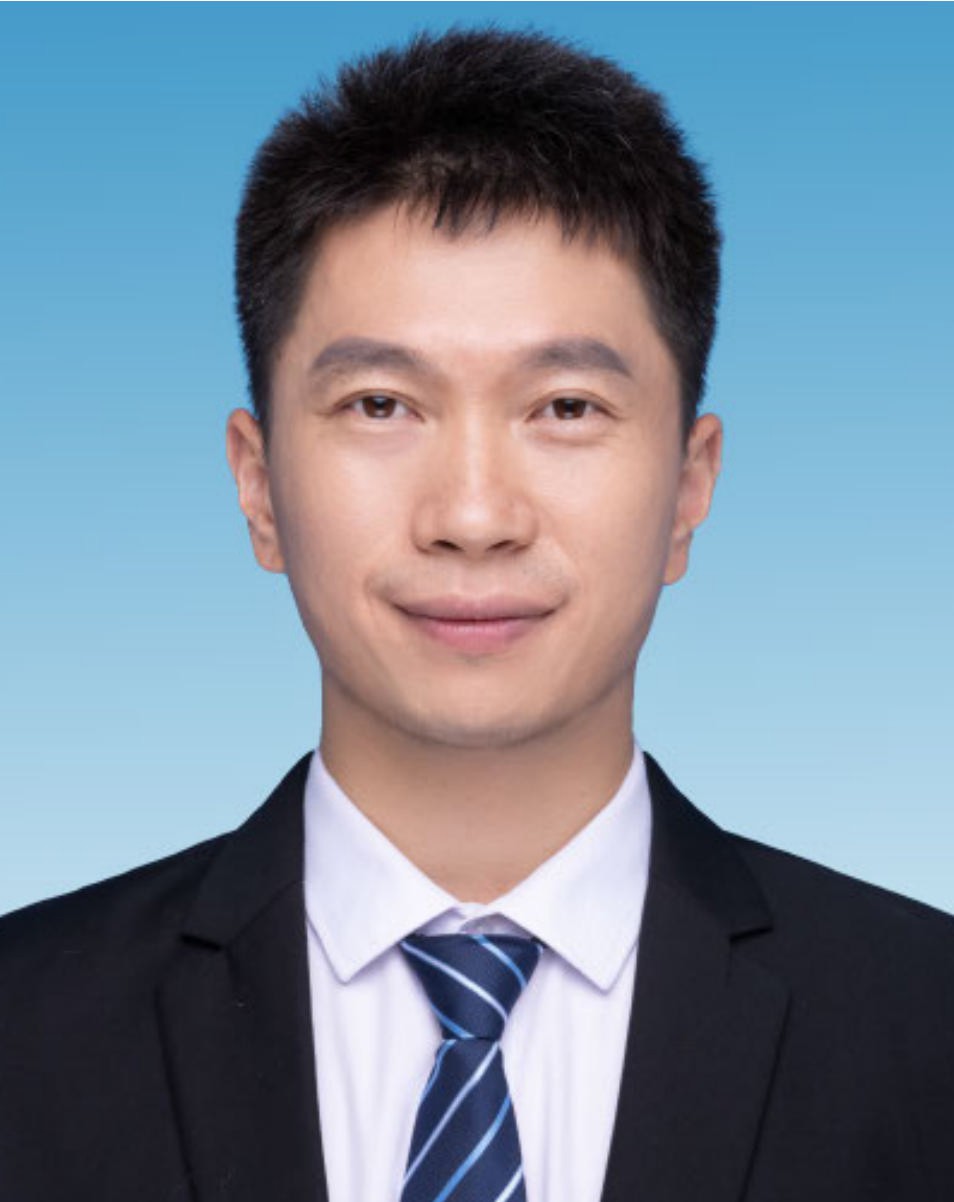}}]{Chenggang Yan}
received the B.S. degree in control science and engineering from Shandong University, Shandong, China, in 2008, and the Ph.D. degree in computer science from the Chinese Academy of Sciences University, Beijing, China, in 2013. He is currently a Professor with the Department of Automation, Hangzhou Dianzi University, Hangzhou, China. His research interests include computational photography and pattern recognition and intelligent system.
\end{IEEEbiography}

\begin{IEEEbiography}[{\includegraphics[width=1in,height=1.25in,clip,keepaspectratio]{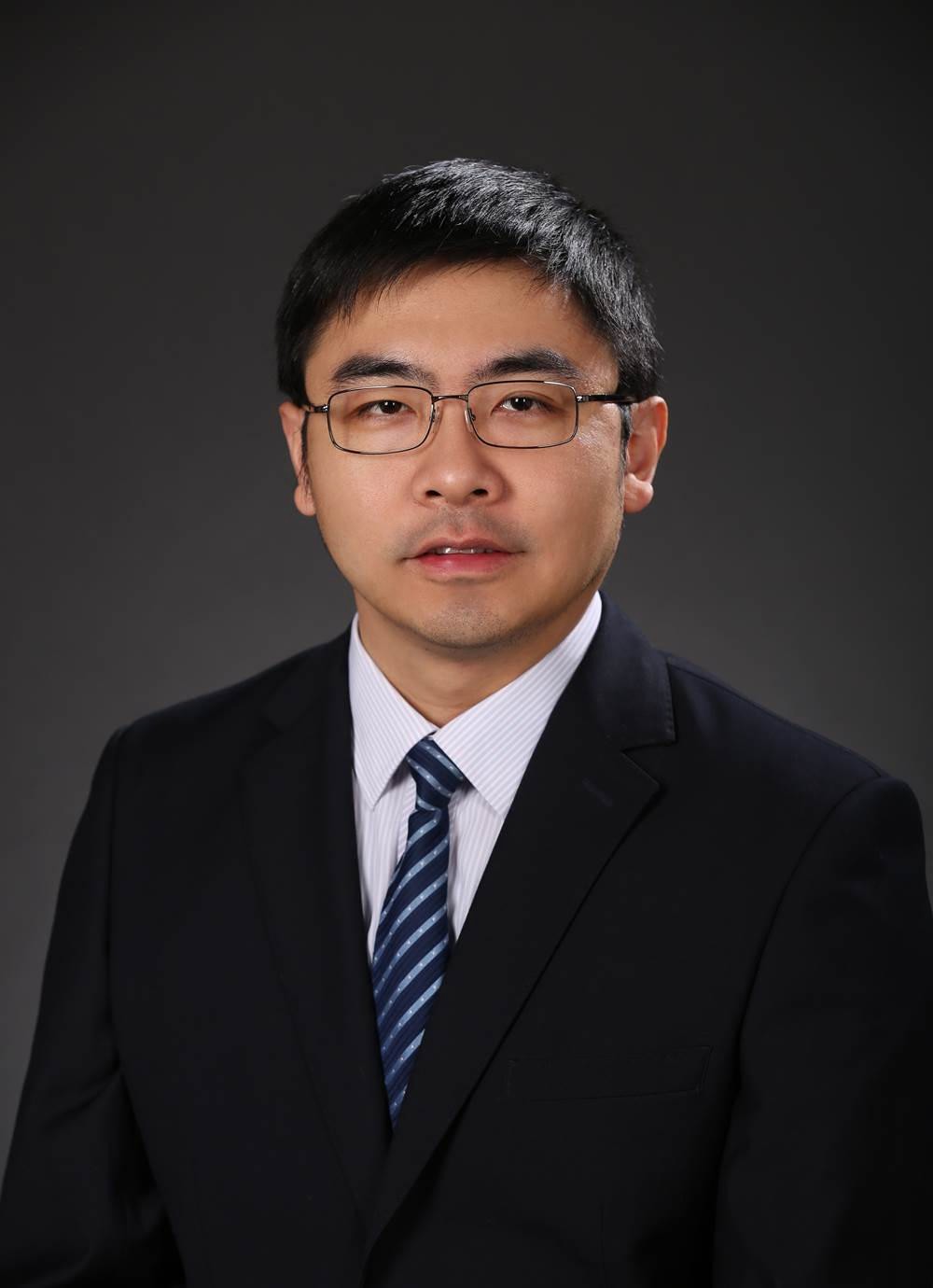}}]{Feng Xu}
received the B.S. degree in physics and the Ph.D. degree in automation from Tsinghua University, Beijing, China, in 2007 and 2012, respectively. He is currently an Associate Professor with the School of Software, Tsinghua University. His research interests include face animation, performance capture, and 3-D reconstruction.
\end{IEEEbiography}

\end{document}